\documentclass{article}

\usepackage[nonatbib,preprint]{neurips_2021}

\usepackage{mysymbols}
\usepackage[utf8]{inputenc} 
\usepackage[T1]{fontenc}    
\usepackage{hyperref}       
\usepackage{url}            
\usepackage{booktabs}       
\usepackage{amsfonts}       
\usepackage{nicefrac}       
\usepackage{microtype}      
\usepackage{xcolor}         
\usepackage{subcaption, booktabs}   
\usepackage{graphicx}
\usepackage{float} 
\usepackage{multirow}
\usepackage{wrapfig}
\usepackage{longtable}

\title{Human-Adversarial Visual Question Answering}

\author{%
    Sasha Sheng$^\ddagger$\thanks{Equal contribution. Correspondence to \href{mailto:advqa@fb.com}{advqa@fb.com}.} \quad Amanpreet Singh$^\ddagger$\footnotemark[1] \quad Vedanuj Goswami$^\ddagger$ \quad Jose Alberto Lopez Magana\thanks{Work done as an intern at Facebook AI Research.} \And \quad Wojciech Galuba$^\ddagger$ \quad Devi Parikh$^\ddagger$ \quad Douwe Kiela$^\ddagger$\\
    $^\ddagger$ Facebook AI Research \quad $^\dagger$ Tecnológico de Monterrey \\
    
}

\newcommand{\acronym}{AdVQA}

\begin{document}

\maketitle

\begin{abstract}

Performance on the most commonly used Visual Question Answering dataset (VQA v2) is starting to approach human accuracy. However, in interacting with state-of-the-art VQA models, it is clear that the problem is far from being solved. In order to stress test VQA models, we benchmark them against human-adversarial examples. Human subjects interact with a state-of-the-art VQA model, and for each image in the dataset, attempt to find a question where the model's predicted answer is incorrect. We find that a wide range of state-of-the-art models perform poorly when evaluated on these examples. We conduct an extensive analysis of the collected adversarial examples and provide guidance on future research directions. We hope that this Adversarial VQA (\acronym{}) benchmark can help drive progress in the field and advance the state of the art.\footnote{We plan to make an evaluation server available for community to evaluate on \acronym{}.}
\end{abstract}

\section{Introduction}

Visual question answering (VQA) is widely recognized as an important evaluation task for vision and language research. Besides direct applications such as helping the visually impaired or multimodal content understanding on the web, it offers a mechanism for probing machine understanding of images via natural language queries. Making progress on VQA requires bringing together different subfields in AI -- combining advances from natural language processing (NLP) and computer vision together with those in multimodal fusion -- making it an exciting task in AI research.

Over the years, the performance of VQA models has started to plateau on the popular VQA v2 dataset~\cite{vqa2} -- approaching inter-human agreement -- as evidenced by Fig.~\ref{fig:vqa_progress}. This raises important questions for the field: To what extent have we solved the problem? If we haven't, what are we still missing? How good are we really?

An intriguing method for investigating these questions is dynamic data collection, where human annotators and state-of-the-art models are put ``in the loop'' together to collect data adversarially. Annotators are tasked with and rewarded for finding model-fooling examples, which are then verified by other humans. The easier it is to find such examples, the worse the model’s performance can be said to be. The collected data can be used to ``stress test'' current VQA models and can serve as the next iteration of the VQA benchmark that can help  drive further progress.

The commonly used VQA dataset \cite{vqa2} dataset was collected by instructing annotators to ``ask a question about this scene that [a]
smart robot probably can not answer'' \cite{vqa1}. One way of thinking about our proposed human-adversarial data collection is that it explicitly ensures that the questions can not be answered by today's ``smartest'' models.

This work is, to the best of our knowledge, the first to apply this human-adversarial approach to any multimodal problem. We introduce Adversarial VQA (\acronym{}), a large evaluation dataset of 28,522 examples in total, all of which fooled the VQA 2020 challenge winner, MoViE+MCAN \cite{nguyen2020revisiting} model.

We evaluate a wide range of existing VQA models on \acronym{} and find that their performance is significantly lower than on the commonly used VQA v2 dataset~\cite{vqa2} (see Table~\ref{tab:teaser}). Furthermore, we conduct an extensive analysis of \acronym{} characteristics, and contrast with the VQA v2 dataset. 

We hope that this new benchmark can help advance the state of the art by shedding important light on the current model shortcomings. Our findings suggest that there is still considerable room for continued improvement, with much more work remaining to be done.

\begin{table*}[t]
    \centering
    \footnotesize
    \setlength{\tabcolsep}{5pt}
    \renewcommand{\arraystretch}{1}
    \caption{\textbf{Contrastive examples from VQA and \acronym{}}. Predictions are given for the VisualBERT, ViLBERT and UniT models, respectively. Models can answer VQA questions accurately, but consistently fail on \acronym{} questions.}
    \begin{tabular}{@{}lp{3cm}p{3cm}@{}}
    \toprule
    \textbf{Image} & \textbf{VQA} & \textbf{AdVQA}\\\midrule
    \raisebox{-.5\height}{\includegraphics[width=3cm]{images/1.jpeg}} &
    \hbox{\textbf{Q}: How many cats are in}
    \hbox{the image?}
    \hbox{\textbf{A}: 2}
    \hbox{\textbf{Model}: 2, 2, 2}
    &
    \hbox{\textbf{Q}: What brand is the tv?}
    \hbox{\textbf{A}: lg}
    \hbox{\textbf{Model}: sony, samsung, samsung}
    \\
    \raisebox{-.5\height}{\includegraphics[width=3cm]{images/2.jpeg}} &
    \hbox{\textbf{Q}: Does the cat look happy?}
    \hbox{\textbf{A}: no}
    \hbox{\textbf{Model}: no, no, no}
    &
    \hbox{\textbf{Q}: How many cartoon drawings}
    \hbox{are present on the cat's tie?}
    \hbox{\textbf{A}: 4}
    \hbox{\textbf{Model}: 1, 1, 2}
    \\
    \raisebox{-.5\height}{\includegraphics[width=3cm]{images/3b.png}} &
    \hbox{\textbf{Q}: What kind of floor is the}
    \hbox{man sitting on?}
    \hbox{\textbf{A}: wood}
    \hbox{\textbf{Model}: wood, wood, wood}
    &
    \hbox{\textbf{Q}: Did someone else take this}
    \hbox{picture?}
    \hbox{\textbf{A}: no}
    \hbox{\textbf{Model}: yes, yes, yes}
    \\
    \bottomrule
    \end{tabular}
    \label{tab:teaser}
\end{table*}

\section{Related Work}

\paragraph{Stress testing VQA.}

Several attempts exist for stress testing VQA models. Some examine to what extent VQA models' predictions are grounded in the image content, as opposed to them relying primarily on language biases learned from the training dataset. The widely used VQA v2 dataset~\cite{vqa2} was one attempt at this. The dataset contains pairs of similar images that have different answers to the same question, rendering a language-based prior inadequate. VQA under changing priors (VQA-CP)~\cite{vqa-cp} is a more stringent test where the linguistic prior not only is weak in the test set, but is adversarial relative to the training dataset (\ie, answers that are popular for a question type during training are rarer in the test set and vice versa). Datasets also exist that benchmark specific capabilities in VQA models, such as reading and reasoning about text in images~\cite{textvqa}, leveraging external knowledge~\cite{okvqa}, and spatial~\cite{gqa} or compositional reasoning~\cite{clevr}. Other vision and language datasets, such as Hateful Memes \cite{hateful}, have also tried to make sure the task involves true multimodal reasoning, as opposed to any of the individual modalities sufficing for arriving at the correct label.

\begin{wrapfigure}{r}{0.53\textwidth}
    \centering
    \includegraphics[width=0.52\columnwidth]{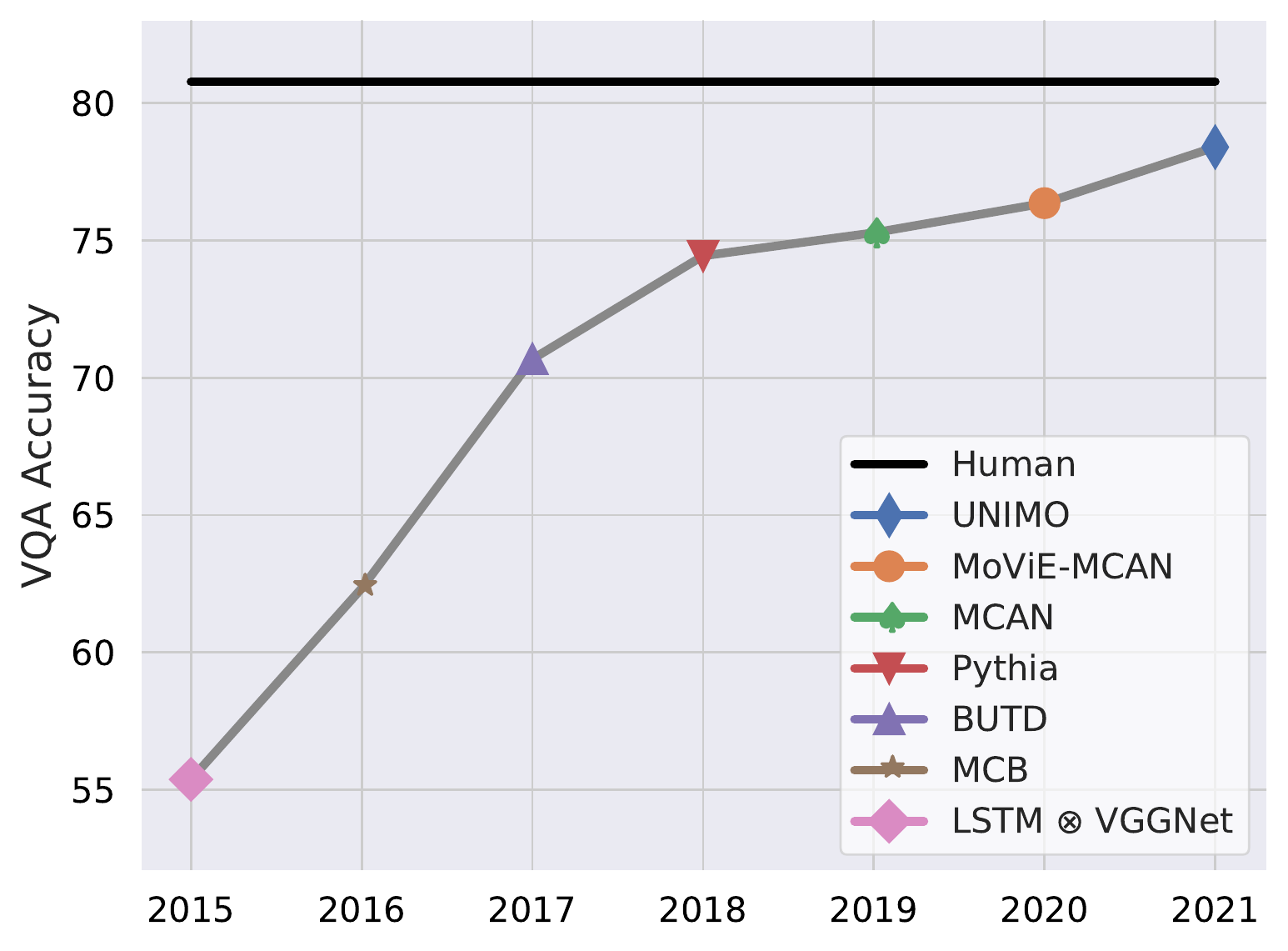}
    \hspace{1mm}
    \caption{Progress on the VQA v2 dataset over time.}
    \label{fig:vqa_progress}
\end{wrapfigure}

\paragraph{Saturating prior work.} The VQA v2 \cite{vqa2} challenge has been running yearly since 2016 and has seen tremendous progress, as can be seen in Figure~\ref{fig:vqa_progress}. From simple LSTM~\cite{hochreiter1997long} and VGGNet~\cite{simonyan2014very} fused models to more advanced fusion techniques (MCB~\cite{fukui2016multimodal}, Pythia~\cite{jiang2018pythia}) to better object detectors~(MCAN~\cite{yu2019deep}, MoViE+MCAN~\cite{nguyen2020revisiting}, BUTD~\cite{anderson2018bottom}) to transformers \cite{vaswani2017attention,singh2020we} and self-supervised pretraining~(MCAN~\cite{yu2019deep}, UNIMO~\cite{li2020unimo}), the community has brought the performance of models on the VQA v2 dataset close to human accuracy, thus starting to saturate progress on dataset. As we show through \acronym{}, however, saturation is far from achieved on the overall task, and hence there is a need for new datasets to continue benchmarking progress in vision and language reasoning. 

\paragraph{Adversarial datasets.}

As AI models are starting to work well enough in some narrow contexts for real world deployment, there are increasing concerns about the robustness of these models. This has led to fertile research both in designing adversarial examples to ``attack'' models (e.g., minor imperceptible noise added to image pixels that significantly changes the model's prediction \cite{goodfellow2014adversarial}), as well as in approaches to make models more robust to ``defend'' against these attacks \cite{carlini2017towards,cisse2017parseval}. The human-in-the-loop adversarial setting that we consider in this paper is qualitatively different from the statistical perturbations typically explored when creating adversarial examples. This type of human-in-the-loop adversarial data collection have been explored in NLP e.g., natural language inference~\cite{nie2020anli}, question answering~\cite{bartolo2020beat,rodriguez2019quizbowl}, sentiment analysis~\cite{potts2020dynasent}, hate speech detection~\cite{vidgen2020learning} and dialogue safety~\cite{dinan2019build}.
To the best of our knowledge, ours is the first work to explore a multimodal human-adversarial benchmark.

The aim of this work is to investigate state of the art VQA model performance via human-adversarial data collection. In this human-and-model-in-the-loop paradigm, human annotators are tasked with finding examples that fool the model. In this case, annotators are shown an image and are tasked with asking difficult but valid questions that the model answers incorrectly. We collected our dataset using Dynabench \cite{kiela2021dyna}, a platform for dynamic adversarial data collection and benchmarking. For details on our labeling user interface, please refer to the supplementary material. 

\begin{figure}[t]
    \centering
    \includegraphics[width=\textwidth]{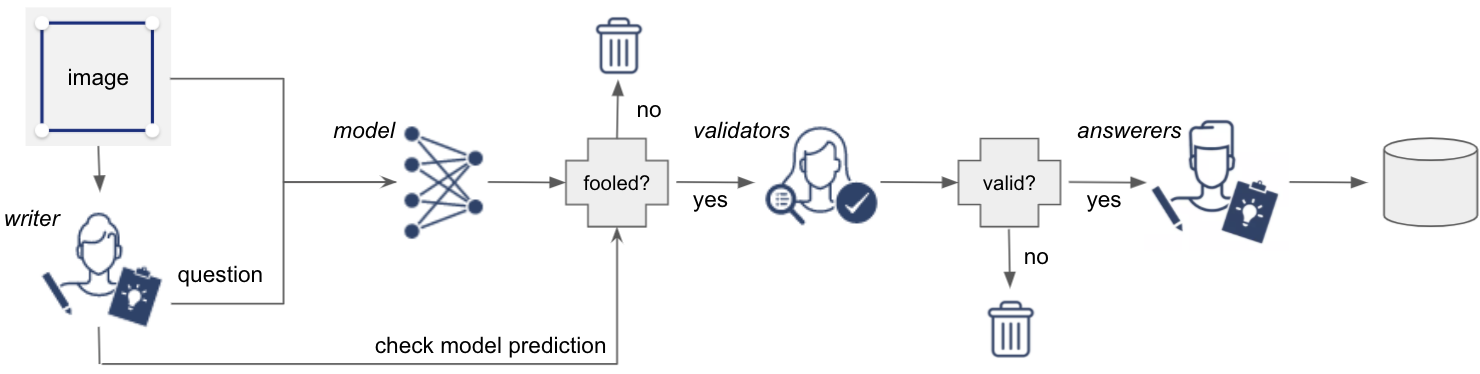}
    \caption{Diagram of the \acronym{} human-adversarial data collection flow.}
    \label{fig:schematic}
\end{figure}

The VQA v2 dataset is based on COCO~\cite{lin2014coco} images.
We collected adversarial questions for subsets of the val2017 COCO images (4,095) and test COCO images (22,259), respectively. The data collection involves three phrases ( Figure~\ref{fig:schematic}): First, in the \textbf{question collection} phase, Amazon Mechanical Turk workers interact with a state-of-the-art VQA model and self-report image-question pairs that the model failed to answer correctly. Second, in the \textbf{question validation} phase, a new set of workers validates whether the answers provided by the VQA model for the image-question pairs collected in the first phrase are indeed incorrect. Finally, in the \textbf{answer collection} phase, we collect 10 ground truth answers for the image-question pairs validated in the second phase. In what follows, we provide further details for each of these steps.

\subsection{Question Collection}

In this phase, annotators on Amazon MTurk are shown an image and are tasked with asking questions about this image. The interface has a state-of-the-art VQA model in the loop. For each question the annotator asks, the model produces an answer. Annotators are asked to come up with questions that fool the model. 
Annotators are done (\ie, can submit the HIT) once they successfully fooled the model or after they tried a minimum of 10 questions (whichever occurs first). To account for the fact that it may be hard to think of many questions for some images, annotators are given the option to skip an image after providing non-fooling three questions. We use the VQA Challenge 2020 winner -- MoViE+MCAN \cite{nguyen2020revisiting} -- trained on the COCO 2017 train set as the model in the loop. This is to ensure that the adversarial benchmark we collect is challenging for the current state-of-the-art in VQA.

\begin{figure*}
    \centering
    \begin{subfigure}[b]{0.53\textwidth}
        \centering
        \includegraphics[width=1.0\columnwidth]{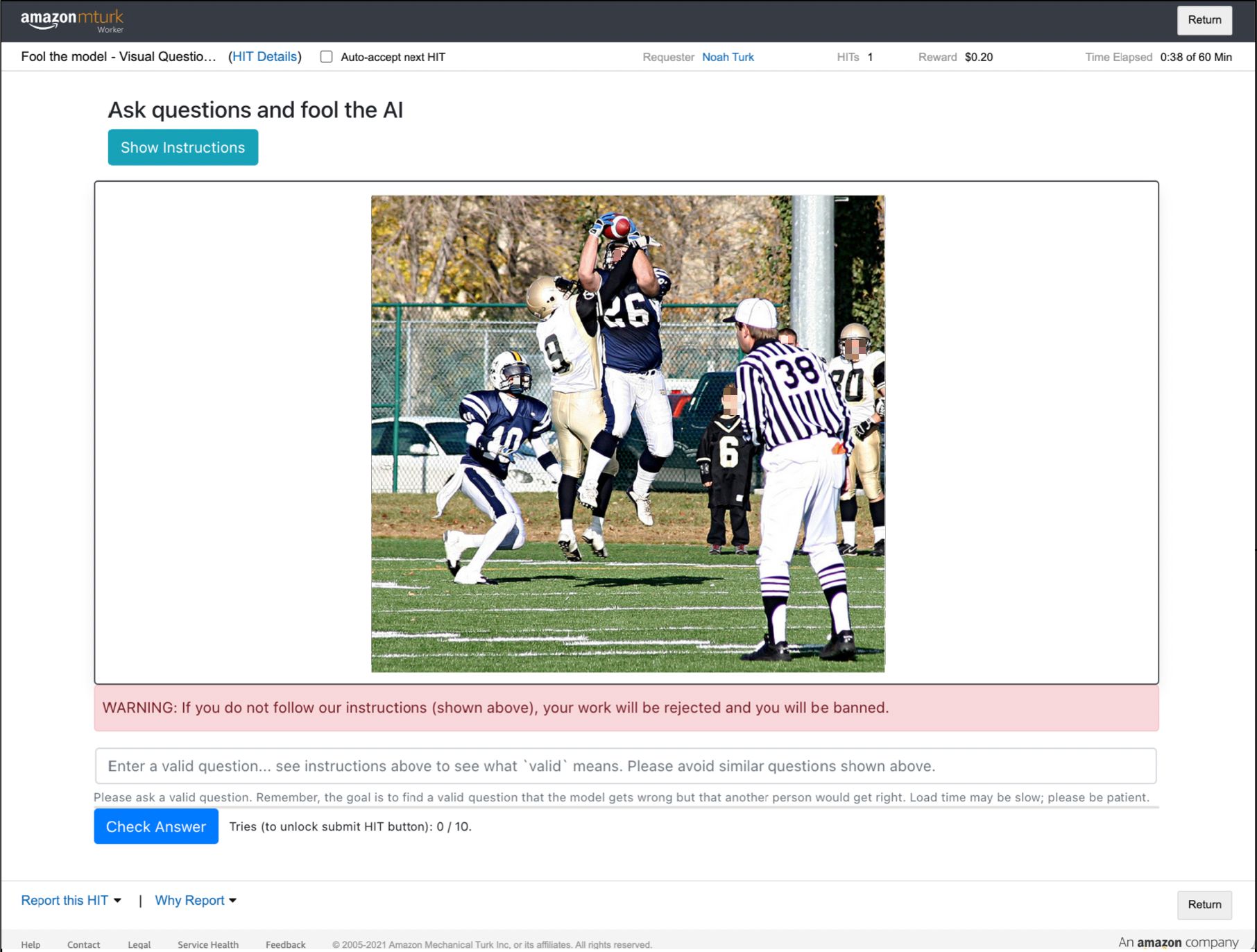}
        \caption{\textbf{Main interface}. The annotators follow the instructions provided to write the question in the text field provided which is then answered by the model-in-the-loop.}
        \label{fig:qc}
    \end{subfigure}
    \hspace{1mm}
    \begin{subfigure}[b]{0.30\textwidth}
        \centering
        \includegraphics[width=0.95\columnwidth]{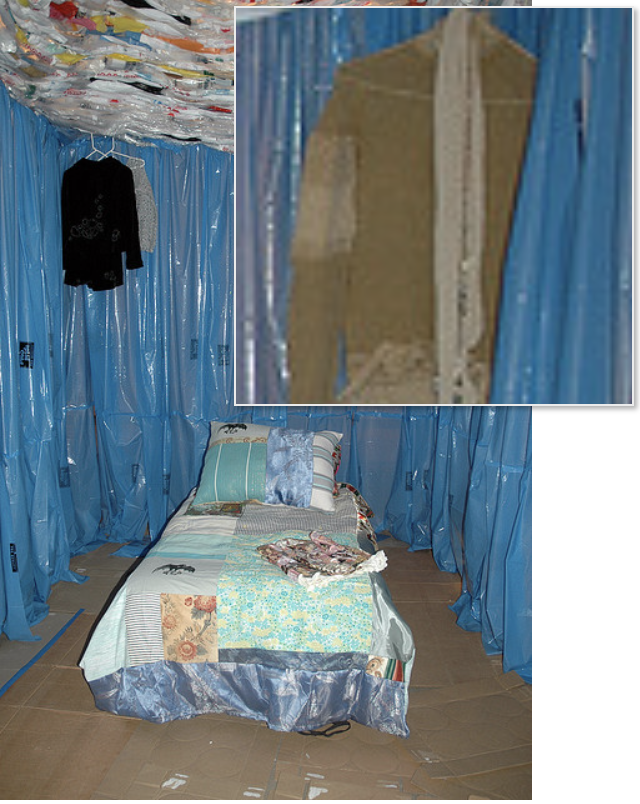}
        \caption{\textbf{Magnifier,} available both in question collection, validation and answer collection helps annotators ask and answer more specific questions.}
        \label{fig:magnifier}
    \end{subfigure}
    \caption{\textbf{Question collection interface} showing the first stage of \acronym{} collection setup.}
    \label{fig:qc_interface}
\end{figure*}

The interface provides a digital magnifier to zoom into an area of the image when the mouse hovers over it, allowing workers to examine the image closely if necessary. See Figure \ref{fig:magnifier}. Keyboard shortcuts were provided for convenience.

\subsection{Question Validation}
\label{sub:validation}
Note that in the question collection phase, annotators self-report when they have identified a question that is incorrectly answered by the model. Whether or not this was actually the case is verified in the question validation phase. Two different annotators are shown the image, question, and answer predicted by the model, and asked whether the model's answer is correct. As an additional quality control, the annotator is also asked whether the question is ``valid''. A valid question is one where the image is necessary and sufficient to answer the question. Examples of invalid questions are: ``What is the capital of USA?'' (does not need an image) or ``What is the person doing?'' when the image does not contain any people or where there are multiple people doing multiple things. If the two validators disagree, a third annotator is used to break the tie. Examples are added to the dataset only if at least two annotators agree that the question is valid.

\subsection{Answer Collection}
In the final stage of data collection, following \cite{vqa2,textvqa}, we collect 10 answers per question providing instructions similar to those used in \cite{vqa2} making sure that no annotator sees the same question twice. In addition, as an extra step of caution, to filter bad questions that might have passed through last two stages and to account for ambiguity that can be present in questions, we allow annotators to select \textit{``unanswerable''} as an answer.
Further, to ensure superior data quality, we occasionally provide annotators with hand-crafted questions for which we know the non-ambiguous single true answer, as a means to identify and filter out annotators providing poor quality responses.

\subsection{Human-Adversarial Annotation Statistics}

The statistics for the first two stages (question collection and validation) are shown in Table \ref{tab:qc}. We find that annotators took about 5 tries and on average around 4 minutes to find a model-fooling example. For computing the model error rate, we can look at the instances where annotators claimed they had fooled the model, and where annotators were verified by other annotators to have fooled the model. The latter has been argued to be a particularly good metric for measuring model performance~\cite{kiela2021dyna}. We also further confirm that the model was indeed fooled by running the model-in-the-loop on a subset of examples in which human agreement was 100\% and found the accuracy to be 0.08\% on val, and 0.1\% on test splits of \acronym{} respectively.

\begin{table}[h]
    \centering
    \caption{\textbf{\acronym{} human-adversarial question collection and dataset statistics.} The model error rate is the percentage of examples where the submitted questions fooled the model (either as claimed during question collection, or after validation). We also report the number of attempts (tries) needed before a validated model-fooling example was found, and how long this took, in seconds.}
    \begin{tabular}{ccccc}
    \toprule
    \textbf{Total} & \multicolumn{2}{c}{\textbf{Model error rate}} & \textbf{Tries} & \textbf{Time in sec} \\
    & claimed & validated & \multicolumn{2}{c}{mean/median per ex.} \\ \midrule
    75,669 & 43.71\% (33,074) & 37.69\% (28,522)  & 5.25/4.0 & 226.17/166.66 \\
    \bottomrule \\
    \end{tabular}
    \label{tab:qc}
\end{table}

Interestingly, the ``claimed'' model error rate, based on self-reported model-fooling questions, is similar to that of text-based adversarial question answering, which was at 44.0\% for RoBERTa and 47.1\% for BERT \cite{bartolo2020beat}. The validated error rate for our task is much higher than e.g. for ANLI \cite{nie2020anli}, which was 9.52\% overall, suggesting that fooling models is a lot easier for VQA than for NLI. It appears to be not too difficult for annotators to find examples that the model fails to predict correctly.

\section{Model Evaluation}
\label{sec:results}

\subsection{Baselines and Methods}
We analyze the performance of several baselines and a wide variety of state-of-the-art VQA models on the \acronym{} dataset. We evaluate the same set of models on VQA v2 dataset as a direct comparison.

\textbf{Prior baselines.} We start by evaluating two prior baselines: answering based on the overall majority answer, or the per answer type majority answer in the validation dataset. We use the same answer types as in~\cite{vqa1}. The overall majority answer in \acronym{} is \textit{no}. The majority answer is \textit{no} for ``yes/no'', \textit{2} for ``numbers'' and \textit{unanswerable} for ``others''. See Section \ref{sec:analysis} for more details on answer types.

\textbf{Unimodal baselines.} Next, we evaluate two unimodal pretrained models: i) ResNet-152~\cite{he2016deep} pretrained on Imagenet~\cite{deng2009imagenet}; and ii) BERT~\cite{devlin2018bert}, both finetuned on the task using the visual (image) or textual (question) modality respectively, while ignoring the other modality. We observe that the unimodal text model performs better than unimodal image for both VQA and \acronym{}.

\textbf{Multimodal methods.} We evaluate two varieties of multimodal models: i) unimodal pretrained and ii) multimodal pretrained. In the unimodal pretrained category we explore MMBT~\cite{kiela2019supervised}, MoViE+MCAN~\cite{nguyen2020revisiting} and UniT~\cite{hu2021unit}. These models are initialized from unimodal pretrained weights: BERT pretraining for MMBT; Imagenet + Visual Genome~\cite{krishna2017visual} detection pretraining for MoViE+MCAN; and Imagenet + COCO~\cite{lin2014microsoft} detection pretraining for the image encoder part and BERT pretraining for the text encoder part in UniT. In the multimodal pretrained category, we explore VisualBERT~\cite{li2019visualbert}, VilBERT~\cite{lu2019vilbert, lu202012}, VilT~\cite{kim2021vilt}, UNITER\cite{chen2020uniter} and VILLA~\cite{gan2020large}. These models are first initialized from pretrained unimodal models and then pretrained on different multimodal datasets on proxy self-supervised/semi-supervised tasks before finetuning on VQA. VisualBERT is pretrained on COCO Captions~\cite{chen2015microsoft}; VilBERT is pretrained on Conceptual Captions~\cite{sharma2018conceptual};  ViLT, UNITER and VILLA models are pretrained on COCO Captions~\cite{chen2015microsoft} + Visual Genome~\cite{krishna2017visual} + Conceptual Captions~\cite{sharma2018conceptual} + SBU Captions~\cite{ordonez2011im2text} datasets. 

We find that multimodal models in general perform better than unimodal models, as we would expect given that both modalities are important for the VQA task. 

\textbf{Multimodal OCR methods.} As we will see in Section~\ref{sec:analysis}, a significant amount of questions in \acronym{} can be answered using scene text. 

We test a state-of-the-art TextVQA \cite{singh2020mmf} model, M4C \cite{hu2020iterative} on \acronym{}. We evaluate two versions: (i) trained on VQA 2.0 dataset, and (ii) trained on TextVQA \cite{singh2020mmf} and STVQA \cite{biten2019scene}. In both cases, we use OCR tokens extracted using the Rosetta OCR system \cite{borisyuk2018rosetta}. We also use the same answer vocabulary used by other models for fair comparison.

\begin{table}[h]
    \centering
    \footnotesize
    \caption{\textbf{Model performance on VQA v2 and \acronym{}} val and test sets. * indicates that this model architecture (but not this model instance) was used in the data collection loop.}
    \begin{tabular}{@{}cl|rr|rr@{}}
        \toprule
         \multicolumn{2}{c|}{\textbf{Model}} & \multicolumn{1}{c}{\textbf{VQA}} & \multicolumn{1}{c|}{\textbf{\acronym{}}} & \multicolumn{1}{c}{\textbf{VQA}} & \multicolumn{1}{c}{\textbf{\acronym{}}} \\
        & & \multicolumn{1}{c}{test-dev} & \multicolumn{1}{c|}{test} & \multicolumn{2}{c}{val} \\\midrule
        \multicolumn{2}{c|}{\emph{Human performance}} & 80.78 & 91.18 & 84.73 & 87.53 \\\midrule
        \multicolumn{2}{c|}{\emph{Majority answer (overall)}} & - & 13.38 & 24.67 & 11.65 \\
        \multicolumn{2}{c|}{\emph{Majority answer (per answer type)}} & - & 27.39 & 31.01 & 29.24 \\\midrule
        Model in loop & MoViE+MCAN \cite{nguyen2020revisiting} & 73.56 & 10.33  & 73.51  & 10.24  \\
        \midrule
        \multirow{2}{*}{Unimodal} & ResNet-152 \cite{he2016deep} & 26.37 & 10.85 & 24.82 & 11.22 \\
        & BERT \cite{devlin2018bert} &  39.47 & 26.9 & 39.40 & 23.81 \\\midrule
        \multirow{3}{*}{\shortstack{Multimodal\\(unimodal pretrain)}} & MoViE+MCAN$^*$ \cite{nguyen2020revisiting} & 71.36  & 26.64  & 71.31  & 26.37  \\
        & MMBT \cite{kiela2019supervised} & 58.00  & 26.70  & 57.32  & 25.78  \\
        & UniT \cite{hu2021unit} & 64.36 & 28.15  & 64.32  & 27.55  \\
        \midrule
        \multirow{6}{*}{\shortstack{Multimodal\\(multimodal pretrain)}} & VisualBERT \cite{li2019visualbert} & 70.37  & 28.70  & 70.05  & 28.03  \\
        & ViLBERT \cite{lu2019vilbert} & 69.42  & 27.36  & 69.27  & 27.36 \\
        & ViLT \cite{kim2021vilt} & 64.52  & 27.11  & 65.43  & 27.19  \\
        & UNITER$_{Base}$ \cite{chen2020uniter} & 71.87 & 25.16 & 70.50 & 25.20 \\
        & UNITER$_{Large}$ \cite{chen2020uniter} & 73.57 & 26.94 & 72.71 & 28.03 \\
        & VILLA$_{Base}$ \cite{gan2020large} & 70.94 & 25.14 & 69.50 & 25.17 \\
        & VILLA$_{Large}$ \cite{gan2020large} & 72.29 & 25.79 & 71.40 &  26.18 \\
        \midrule
        \multirow{2}{*}{\shortstack{Multimodal\\(unimodal pretrain + OCR)}} & M4C (TextVQA+STVQA) \cite{hu2020iterative} & 32.89  & 28.86  & 31.44  & 29.08  \\
        & M4C (VQA v2 train set) \cite{hu2020iterative} & 67.66  & 33.52  & 66.21  & 33.33 \\
        
        \bottomrule 
    \end{tabular}
    \label{tab:result_a}
\end{table}

\subsection{Discussion} 
Our model evaluation results show some surprising findings. We discuss our observations and hypotheses around these findings in this section.

\textbf{Baseline comparison.} Surprisingly, most multimodal models are unable to outperform a unimodal baseline (BERT \cite{devlin2018bert}) and simple majority answer (per answer type) prior baseline which only predicts the most-frequently occurring word for the question's category (\textit{no} for ``yes/no'', \textit{2} for ``numbers'' and \textit{unanswerable} for ``others''). A detailed breakdown of category-wise performance provided in Table~\ref{tab:category_wise_performance} suggests that even though ``yes/no'' questions are often considered to be easy, the baseline multimodal models we evaluate are unable to beat the majority answer prediction. We also observe varied but close to majority answer performance in the ``numbers'' category, which is a question type known to be difficult for VQA models \cite{nguyen2020revisiting}. In contrast, the models outperform the majority answer in the ``others'' category even though ``unanswerable'' (the majority answer) is not in the answer vocabulary used. Interestingly, M4C outperforms all on ``numbers'' and ``others'' categories possibly thanks to its text reading capabilities. These trends showcase the difficulty of \acronym{} and suggest that we have a long way to go still, given that we are as yet apparently unable to beat such simple baselines. 

\textbf{Model rankings.} First, M4C \cite{hu2020iterative} performs the best among the evaluated models. Interestingly, it is  smaller than many of the more sophisticated model architectures that score higher on VQA. This is probably due to the importance of the ability to read and reason about text in the image for answering some \acronym{} questions. Second, among models that can't read text, VisualBERT \cite{li2019visualbert} is the best model, despite (or perhaps because of?) it being simple and having fewer parameters compared to other models. Third, the adversarially-trained VILLA \cite{gan2020large} model performs surprisingly poorly. While it may be more robust to statistically-generated adversarial examples, it appears to be less so against human-adversarial examples. Fourth, we find that all of these models perform poorly compared to humans, while model performance on VQA is much closer to that of humans.

\begin{wraptable}{r}{0.5\linewidth}
    \footnotesize
    \setlength{\tabcolsep}{3.2pt}
    \centering
    \caption{\textbf{The category-wise performance of VQA models.} The state-of-the-art VQA models perform very close to the majority class prior, illustrating the challenge and difficulty of \acronym{}.}
    \begin{tabular}{@{}l|r|r|r@{}}
        \toprule
        \multirow[c]{2}{*}{\textbf{Model}} & \multicolumn{3}{c}{\textbf{Question Type}} \\
        & \multicolumn{1}{c|}{yes/no} & \multicolumn{1}{c|}{numbers} & \multicolumn{1}{c}{others} \\\midrule
        Majority Class & 62.28 & 31.11 & 9.29\\\midrule
        ResNet-152 & 62.81 & 0.18 & 0.51 \\\midrule
        BERT & 67.58 & 26.87 & 9.25 \\\midrule
        VisualBERT & 55.51 & 32.29 & 17.66\\\midrule
        ViLBERT & 55.58 & 29.49 & 16.67 \\\midrule
        MoViE+MCAN$^*$ & 52.74 & 33.62 & 14.56 \\\midrule
        M4C (VQA2) & 56.67 & 38.04 & 22.73 \\
        \bottomrule 
    \end{tabular}
    \label{tab:category_wise_performance}
\end{wraptable}

\textbf{Human performance.} Another surprising finding is that inter-human agreement is higher on the \acronym{} dataset than on VQA. This could be due to different data annotation procedures, requirements on annotators or just statistical noise. Human-adversarial questions may also be more specific due to annotators having to make crisp decisions about the model failing or not. 

\textbf{Model-in-the-loop's performance.} Interestingly, the MoViE+MCAN model that was \emph{not} used in the loop and trained with a different seed, performs very similarly to other models. This suggests that to some extent, annotators overfit to the model instance. An alternative explanation is that model selection for all evaluated models was done on the \acronym{} validation set, which was (obviously) not possible for the model in the loop used to construct the dataset. In Adversarial NLI \cite{nie2020anli}, the entire model class of the in-the-loop model was affected. Note however that all VQA models perform poorly on \acronym{}, suggesting that the examples are by and large representative of shortcomings of VQA techniques overall, and not of an individual model instance or class.

\textbf{Train vs Test Distribution.} We experiment with finetuning VisualBERT and VilBERT further on the \acronym{} \textit{val} set, finding that this improved test accuracy from $28.70$ to $33.10$ and $27.36$ to $32.85$, respectively, suggesting a difference in the VQA and \acronym{} distributions, as we would expect. 

\subsection{Training Details}
\label{sec:implementation_details}

We train all the models in Table~\ref{tab:result_a} on the VQA \textit{train} + \textit{val} split excluding the COCO 2017 validation images. We also add questions from Visual Genome~\cite{krishna2017visual} corresponding to the images that overlap with the VQA training set to our training split. The VQA \textit{val} set in our results contains all questions associated with the images in the COCO 2017 validation split. This is also consistent with the training split used for the model in the loop. We collect our \acronym{} validation set on the images from the COCO 2017 \textit{val} split, ensuring there is no overlap with the training set images. We choose the best checkpoint for each model by validating on the \acronym{} validation set. 

For all our experiments, we use the standard architecture for the models as provided by their authors. For the models that are initialized from pretrained models (whether unimodal or multimodal pretrained), we use  off-the-shelf pretrained model weights and then finetune on our training set. We do not do any hyperparamter search for these models and use the best hyperparams as provided by respective authors. We finetune each model with three different seeds and report average accuracy.

We run most of our experiments on NVIDIA V100 GPUs. The maximum number of GPUs used for training is 8 for larger models. Maximum training time is 2 days. More details about hyperparameters, number of devices and time for training each model are provided in the supplementary material.

\begin{figure*}
    \centering
        \begin{subfigure}[b]{0.32\linewidth}
        \includegraphics[width=1\linewidth]{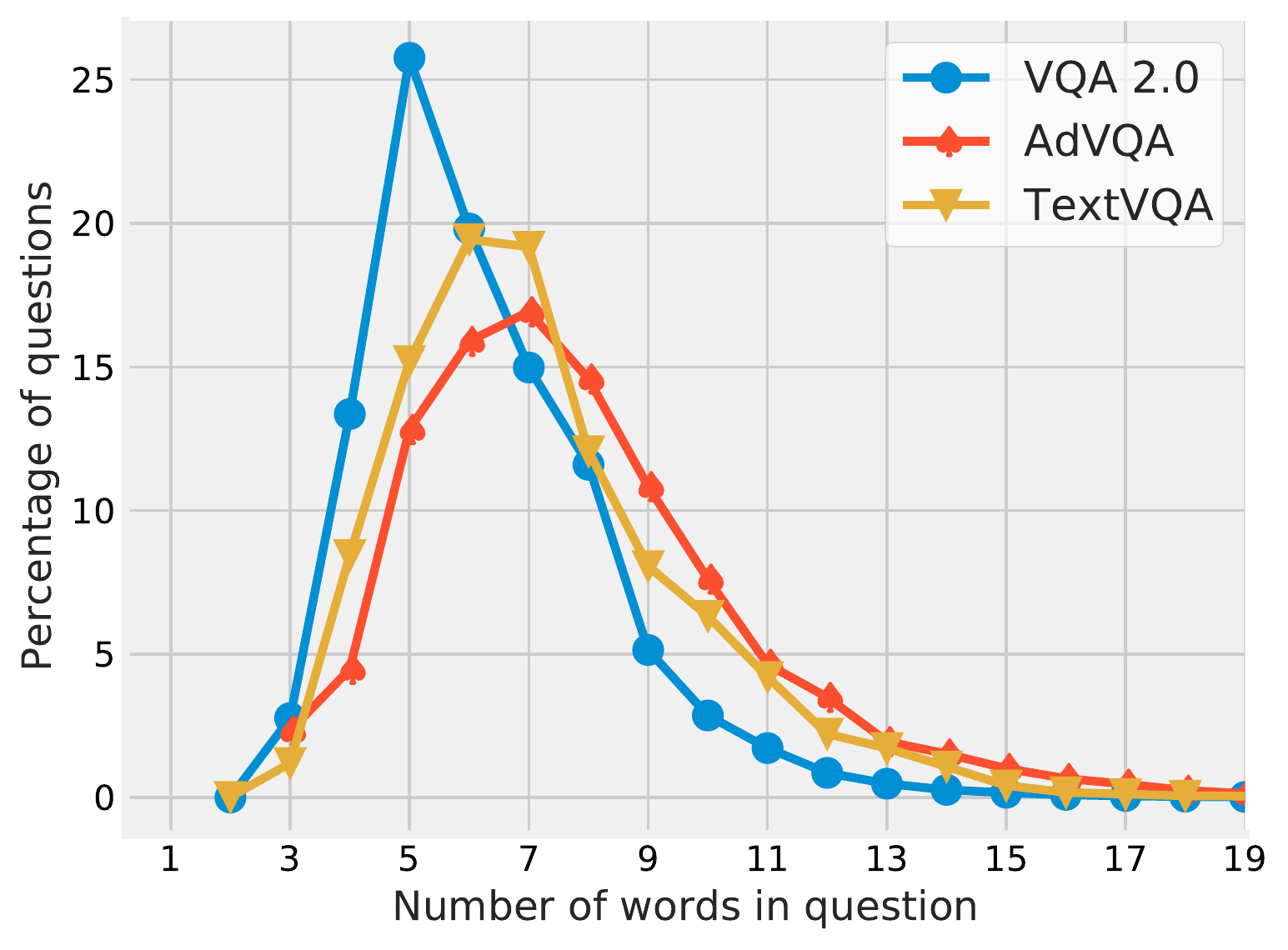}
        \caption{}
        \label{fig:question_lengths}
    \end{subfigure}
    \begin{subfigure}[b]{0.32\linewidth}
        \includegraphics[width=1\linewidth]{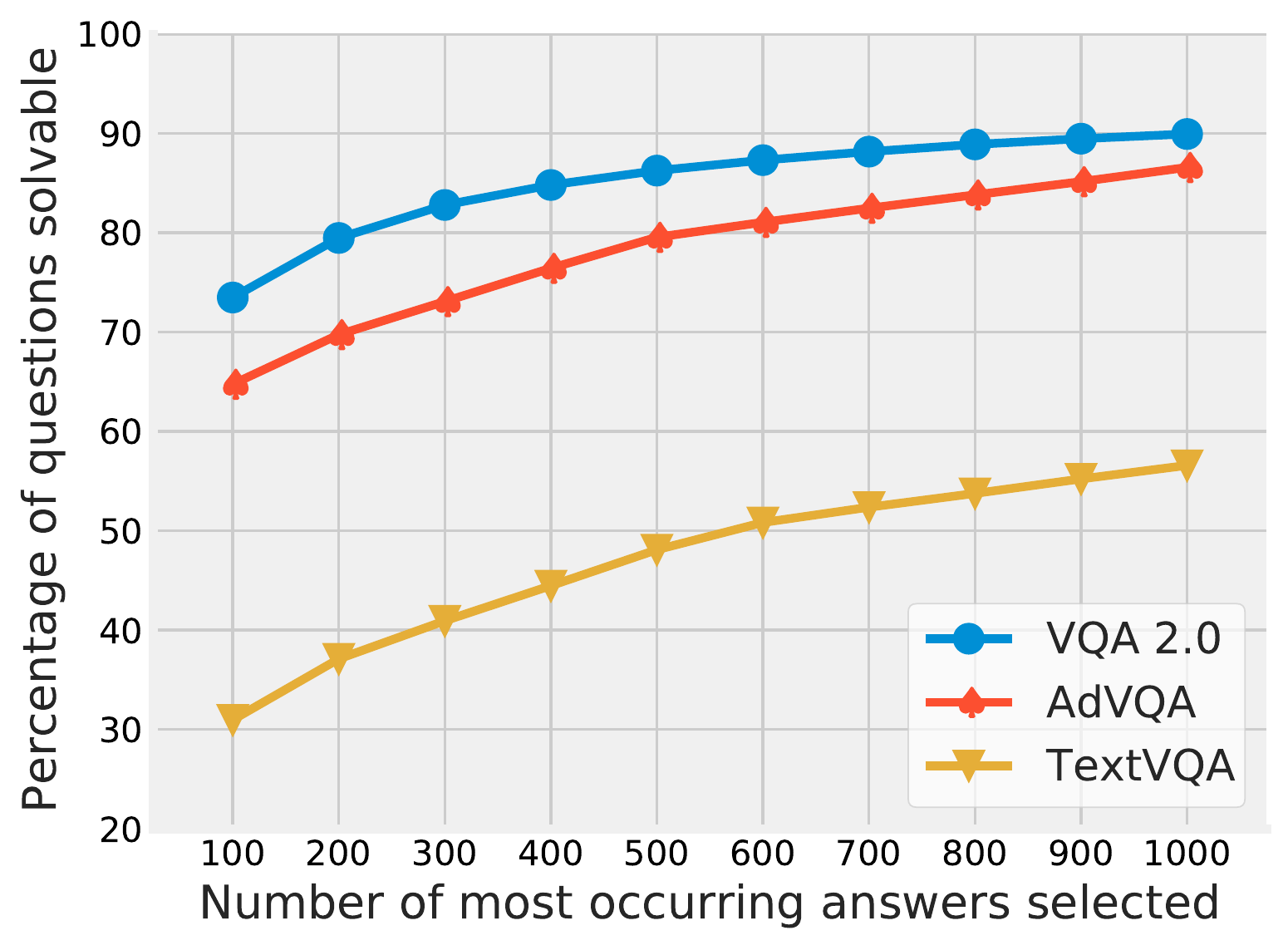}
        \caption{}
        \label{fig:topk_answerable}
    \end{subfigure}
    \begin{subfigure}[b]{0.32\linewidth}
        \includegraphics[width=1\linewidth]{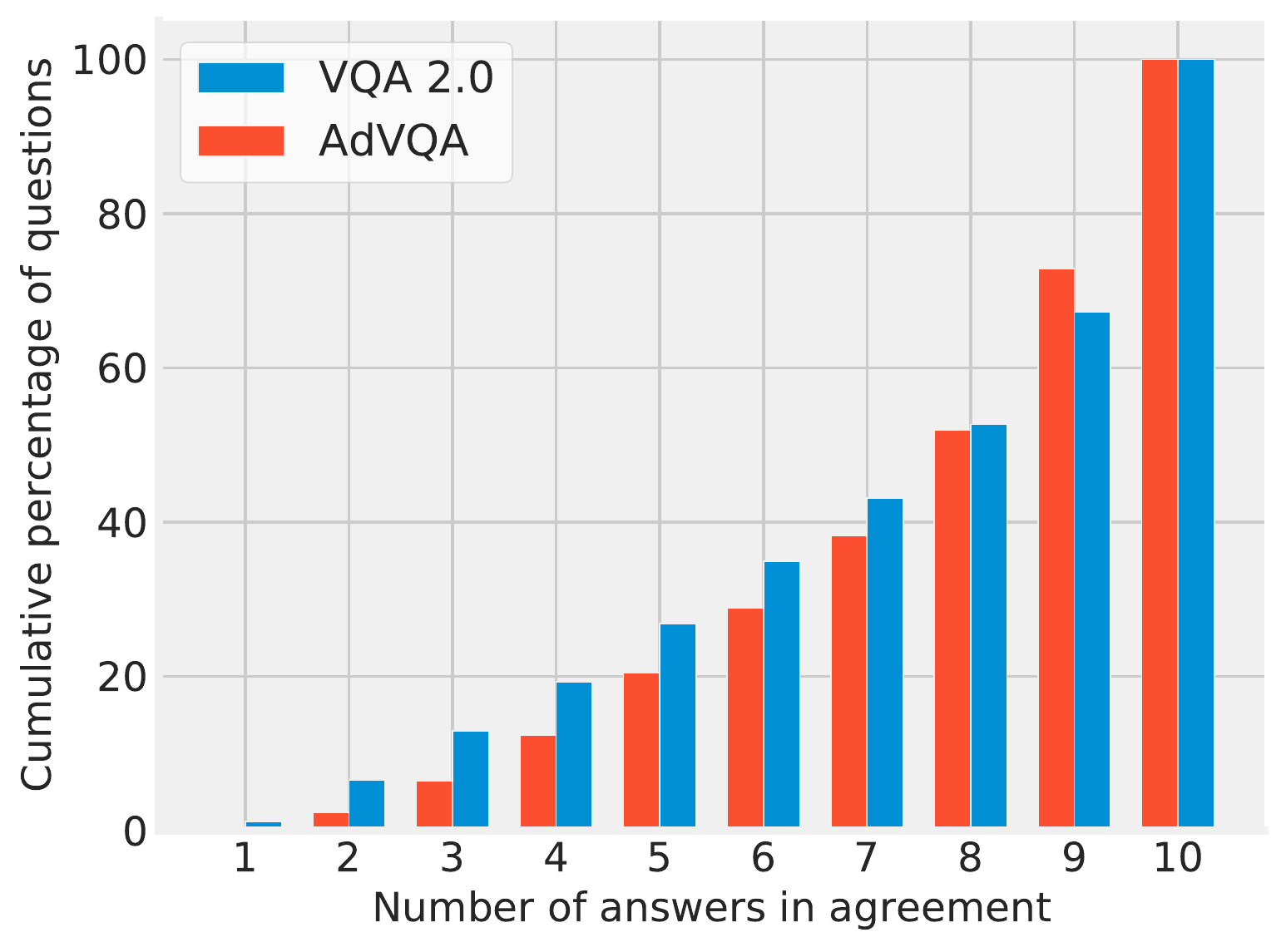}
        \caption{}
        \label{fig:human_agreement}
    \end{subfigure}
    \caption{\textbf{Quantitative statistics for \acronym{} val set questions and answers} showing longer question length, more diversity and better human-agreement. (a) Percentage of questions solvable with a particular question length in \acronym{} val set. 
        We see that the average question length (7.82) in \acronym{} is higher than the prior work. (b) Percentage of questions solvable when a particular number of top k most occurring answers are selected from each dataset. The plot suggests that \acronym{} has a much more diverse answer vocabulary compared to \cite{vqa2} but also not quite as challenging as  \cite{textvqa}. (c) Cumulative human agreement scores. We see that human agreement is better and higher on \acronym{} compared to \cite{vqa2}.}
\end{figure*}

\section{Dataset Analysis}
\label{sec:analysis}
\textbf{Questions.} We first analyze the question diversity in \acronym{} and compare them with popular VQA datasets. \acronym{} contains 28,522 questions~(5,123 in val and 23,399 in test) each with 10 answers.

\begin{wrapfigure}{r}{0.5\textwidth}
    \hspace{-15mm}
    \includegraphics[width=1.25\linewidth]{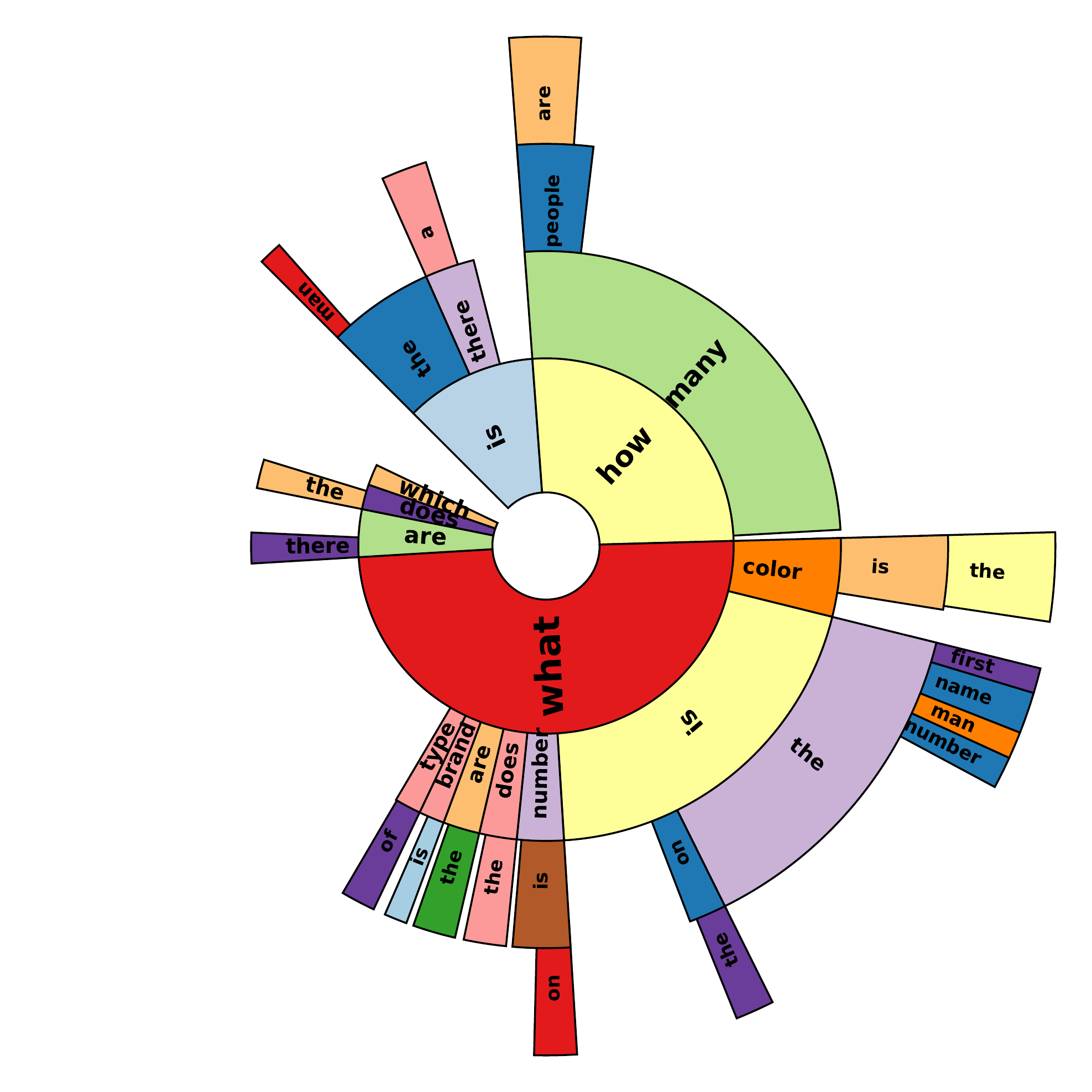}
    \caption{\textbf{Sunburst distribution for the first four words in the \acronym{} val set questions.} Most questions start with ``what'' or ``how''.}
    \label{fig:sunburst}
\end{wrapfigure}

\begin{figure}[t]
    \centering
    \begin{subfigure}[b]{0.45\linewidth}
        \includegraphics[width=1\linewidth]{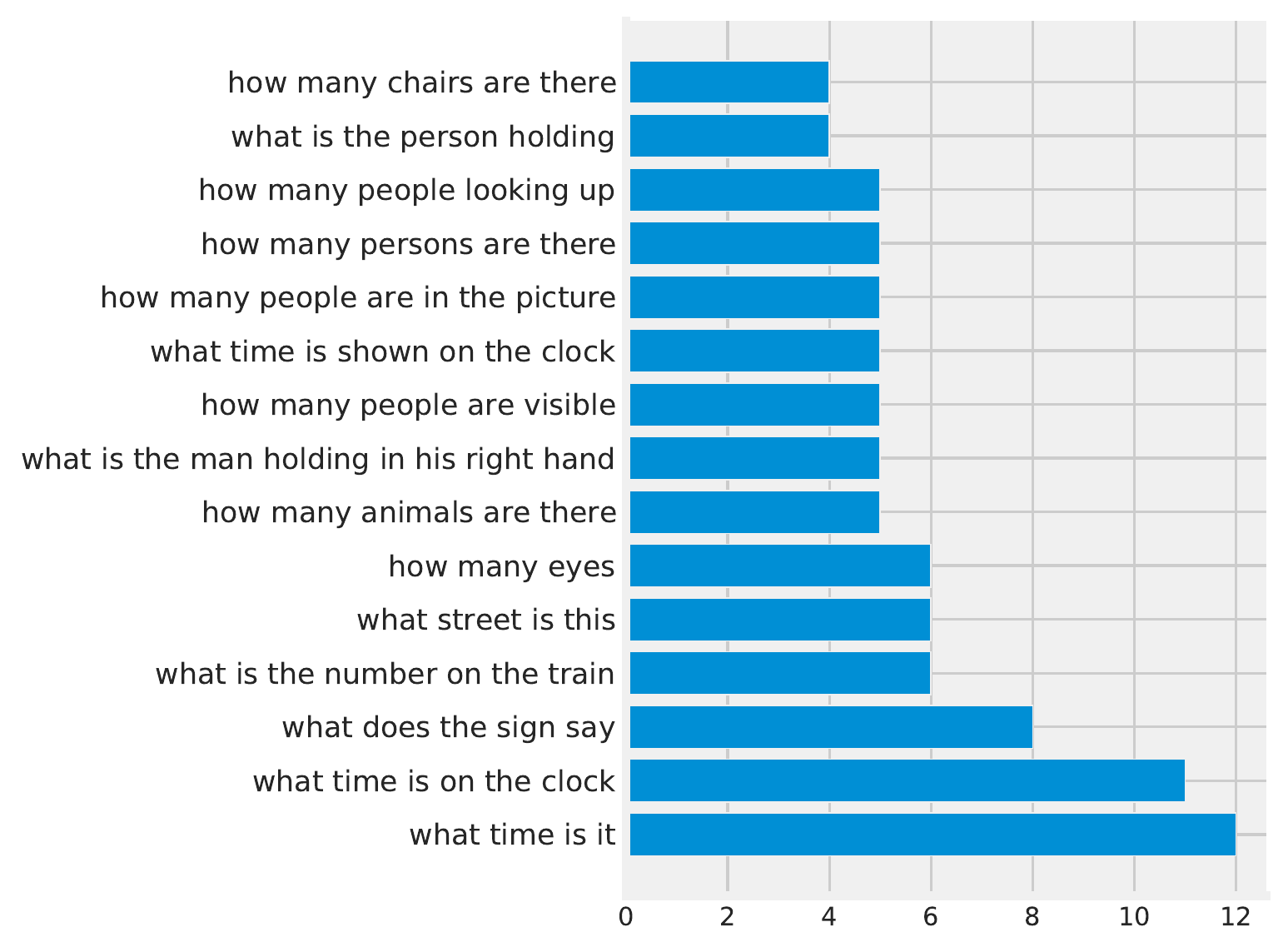}
        \caption{}
        \label{fig:top_questions}
    \end{subfigure}
    \begin{subfigure}[b]{0.45\linewidth}
        \includegraphics[width=1\linewidth]{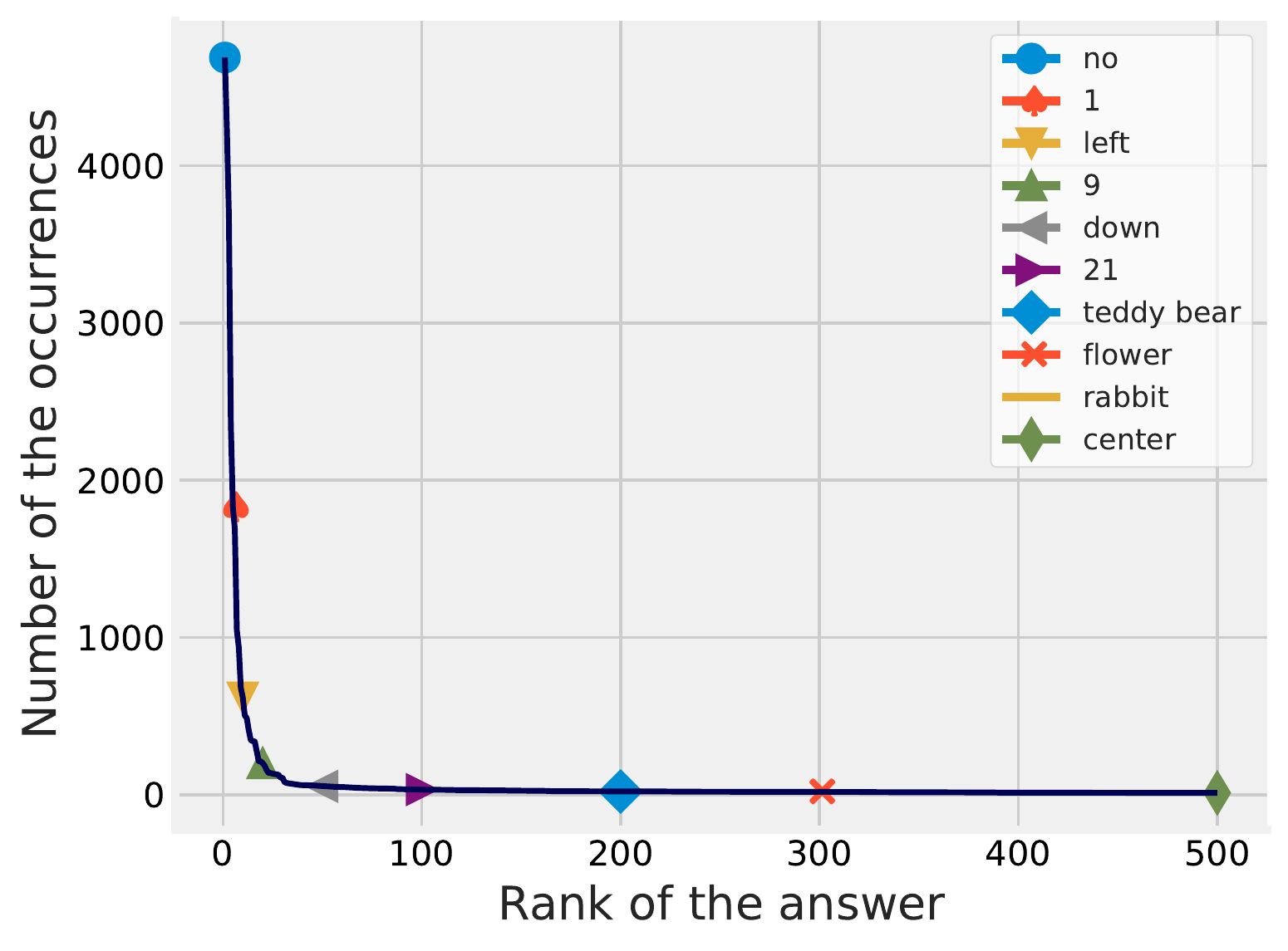}
        \caption{}
        \label{fig:top_answers}
    \end{subfigure}
    \caption{\textbf{Qualitative statistics for \acronym{} val set questions and answers} showing top questions, answers and word distribution. (a) \textbf{Top 15 most occurring questions in \acronym{} val.} Most of the top questions start with ``what''. (b) \textbf{Total occurrences for 500 most common answers
        } 
        with markers for particular ranks.}
\end{figure}

Fig.~\ref{fig:question_lengths} shows the distribution of question length compared with \cite{vqa2,textvqa}. The average question length in \acronym{} is 7.8, which is higher than the VQA v2 (6.3), and TextVQA (7.2). The workers often need to get creative and more specific to fool the model-in-the-loop, leading to somewhat longer questions (\eg specifying a particular person to ask about). Fig.~\ref{fig:top_questions} shows the top 15 most occurring questions from the val set, showcasing that questions involving text (\eg time, sign) and counting (\eg how many) are major failures for current state-of-the-art VQA models, corroborating the findings of prior work in \cite{textvqa,gurari2018vizwiz,trott2017interpretable,clevr}. Fig.~\ref{fig:sunburst} shows a sunburst plot for the first 4 words in the \acronym{} val set questions. We can observe that questions in \acronym{} often start with ``what'' or ``how'' frequently inquiring about aspects like ``many'' (count) and ``brand'' (text).

\textbf{Answers.} In \acronym{} val set, only 66.6\% (3,856)  answers occur more than twice compared to 94.8\% in VQA v2 \cite{vqa2}, suggesting that the diversity of possible answers is much larger in \acronym{} compared to VQA v2. Fig.~\ref{fig:topk_answerable} shows the percentage of questions that are solvable with a vocabulary of top k most occurring answers. We observe that more questions in VQA v2 while fewer question in TextVQA are solvable with smaller vocabulary compare to \acronym{}. This suggests that \acronym{} is more diverse and difficult than VQA v2 but not as narrowly focused as TextVQA, making it a great testbed for future VQA models. We also showcase more qualitative examples in our supplementary to demonstrate that \acronym{}'s diversity doesn't lead to unnatural questions. Fig.~\ref{fig:human_agreement} shows the cumulative percentage of questions where more than a particular number of annotators agree with each other. Fig.~\ref{fig:top_answers} shows the top 500 most occurring answers in the \acronym{} val set; starting from very common answers such as ``no'' and counts (``1'', ``9''), to gradually more specific and targeted answers like ``21'', ``teddy bear'' and ``center''. 

\textbf{Answer Vocabulary.} To showcase the challenge of \acronym{}, we take the original VQA v2 vocabulary used in the Pythia v0.3 model \cite{singh2020mmf}. We find that 77.2\% of the \acronym{} val set's questions are answerable using this vocabulary, suggesting that a model with powerful reasoning capability won't be heavily limited by vocabulary on \acronym{}. But, we also note that for high performance on \acronym{}, a model will need to understand and reason about rare concepts, as 50.9\% of the answers in \acronym{} val and test sets don't occur in VQA v2 train set.

\begin{wraptable}{r}{0.5\textwidth}
    \footnotesize
    \setlength{\tabcolsep}{3.2pt}
    \renewcommand{\arraystretch}{1}
    \centering
    \caption{\textbf{The category-wise distribution of answers.} Compared to VQA, \acronym{} contains more ``number''based and lesser ``yes/no'' questions supporting the prior work's observations around failure of VQA models to count and read text.}
    \begin{tabular}{@{}l|rr|rr@{}}
        \toprule
        \multirow[c]{2}{*}{\textbf{\shortstack{Question \\ Type}}} & \multicolumn{1}{c}{\textbf{VQA}} & \multicolumn{1}{c|}{\textbf{\acronym{}}} & \multicolumn{1}{c}{\textbf{VQA}} & \multicolumn{1}{c}{\textbf{\acronym{}}} \\
        & \multicolumn{1}{c}{test-dev} & \multicolumn{1}{c|}{test} & \multicolumn{2}{c}{val} \\\midrule
        yes/no & 38.36 & 17.89 & 37.70 & 17.90\\\midrule
        number & 12.31 & 41.91 & 11.48 & 31.80 \\\midrule
        others & 49.33 & 40.20 & 50.82 & 50.30 \\
        \bottomrule 
    \end{tabular}
    \label{tab:category_wise_distribution}
\end{wraptable}

\textbf{Question Types.} Table~\ref{tab:category_wise_distribution} shows the category-wise distribution for \acronym{} questions compared with VQA v2 \cite{vqa2}. We can observe a shift from more easy questions of the ``yes/no'' category in VQA v2 dataset to more difficult questions in ``numbers'' category (as suggested in prior work \cite{nguyen2020revisiting}) in \acronym{}.

\textbf{Human Agreement.} In the \acronym{} val set, 27.2\% human annotators agree on all 10 answers while 3 or more annotators agree an answer for 97.7\%, which is higher compared to 93.4\% on VQA v2 \cite{vqa2}, even though as discussed \acronym{} contains a large number of rare concepts. 

\textbf{Relationship to TextVQA \cite{singh2020mmf}.} To understand if the ability to read text is crucial for \acronym{}, we extract Rosetta \cite{borisyuk2018rosetta} tokens on the \acronym{} val set and determine how many questions can be answered using OCR tokens at an edit distance of 1 to account for OCR errors. We find that 15.4\% of the \acronym{} val questions are solvable using OCR tokens suggesting that ability to read scene text would play a crucial role in solving \acronym{}. 

\section{Conclusion, Limitations \& Outlook}

We introduced the \acronym{} dataset, a novel human-adversarial multimodal dataset designed for accelerating progress on Visual Question Answering (VQA). Current VQA datasets have started plateauing and are approaching saturation with respect to human performance. In this work, we demonstrate that the problem is far from solved.

In particular, our analysis and model evaluation results suggest that current state-of-the-art models underperform on \acronym{} due to a reasoning gap incurred from a combination of (i) inability to read text; (ii) inability to count; (iii) heavy bias towards the VQA v2 question and answer distribution; (iv) external knowledge; (v) rare unseen concepts; and (vi) weak multimodal understanding. We've shown the gap is unlikely to be due to (i) limited answer vocabulary; (ii) language representation (BERT performance compared to other); (iii) no pretraining (UniT); or (iv) lack of adversarial training (VILLA performance).
We will provide an evaluation benchmark for community to evaluate on \acronym{} and hope that \acronym{} will help bridge the gap by serving as a dynamic new benchmark for visual reasoning with a large amount of headroom for further progress in the field.

In future work, it would be interesting to continue \acronym{} as a dynamic benchmark. If a new state of the art emerges, those models can be put in the loop to examine how we can improve even further. 

\section*{Broader Impact}

This work analyzed state-of-the-art Visual Question Answering (VQA) models via a dataset constructed using a dynamic human-adversarial approach. We hope that this work can help make VQA models more robust.

VQA datasets contain biases, both in the distribution of images represented in the datasets, as well as the corresponding questions and answers. These biases are likely amplified by the VQA models trained on these datasets. Biases studied in the context of image captioning~\cite{Burns2018WomenAS,biasemnlp} are also relevant for VQA. English is the only language represented in this work.

VQA models can be useful for aiding visually impaired users. The commonly used VQA datasets are not representative of the needs of visually impaired users -- both in terms of the distribution of images (typically consumer photographs from the web), and in terms of the questions contained in the datasets (typically asked by sighted individuals while looking at the image). In contrast, the VizWiz dataset~\cite{gurari2018vizwiz} contains questions asked by visually impaired users on images taken by them to accomplish day-to-day tasks. It would be beneficial for the community to collect larger datasets in this context to enable progress towards relevant technology. It would be important to do this with input from all relevant stakeholders, and in a responsible and privacy-preserving manner. VQA models are currently far from being accurate enough to be uesful or safe in these contexts.

\begin{ack}
We thank our collaborators in the Dynabench team for their support and discussions.
\end{ack}

\bibliographystyle{plain}
\bibliography{arxiv}

\begin{thebibliography}{10}

\bibitem{vqa-cp}
Aishwarya Agrawal, Dhruv Batra, Devi Parikh, and Aniruddha Kembhavi.
\newblock Don't just assume; look and answer: Overcoming priors for visual
  question answering.
\newblock In {\em Proceedings of IEEE/CVF Conference on Computer Vision and
  Pattern Recognition}, 2018.

\bibitem{anderson2018bottom}
Peter Anderson, Xiaodong He, Chris Buehler, Damien Teney, Mark Johnson, Stephen
  Gould, and Lei Zhang.
\newblock Bottom-up and top-down attention for image captioning and visual
  question answering.
\newblock In {\em Proceedings of the IEEE conference on computer vision and
  pattern recognition}, pages 6077--6086, 2018.

\bibitem{vqa1}
Stanislaw Antol, Aishwarya Agrawal, Jiasen Lu, Margaret Mitchell, Dhruv Batra,
  C.~Lawrence Zitnick, and Devi Parikh.
\newblock {VQA}: {V}isual {Q}uestion {A}nswering.
\newblock In {\em Proceedings of IEEE/CVF International Conference on Computer
  Vision}, 2015.

\bibitem{bartolo2020beat}
Max Bartolo, Alastair Roberts, Johannes Welbl, Sebastian Riedel, and Pontus
  Stenetorp.
\newblock Beat the ai: Investigating adversarial human annotation for reading
  comprehension.
\newblock {\em Transactions of the Association for Computational Linguistics},
  8:662--678, 2020.

\bibitem{biten2019scene}
Ali~Furkan Biten, Ruben Tito, Andres Mafla, Lluis Gomez, Mar{\c{c}}al Rusinol,
  Ernest Valveny, CV~Jawahar, and Dimosthenis Karatzas.
\newblock Scene text visual question answering.
\newblock In {\em Proceedings of IEEE/CVF Conference on Computer Vision and
  Pattern Recognition}, pages 4291--4301, 2019.

\bibitem{borisyuk2018rosetta}
Fedor Borisyuk, Albert Gordo, and Viswanath Sivakumar.
\newblock Rosetta: Large scale system for text detection and recognition in
  images.
\newblock In {\em Proceedings of the 24th ACM SIGKDD International Conference
  on Knowledge Discovery \& Data Mining}, pages 71--79, 2018.

\bibitem{Burns2018WomenAS}
Kaylee Burns, Lisa~Anne Hendricks, Trevor Darrell, and Anna Rohrbach.
\newblock Women also snowboard: Overcoming bias in captioning models.
\newblock In {\em Proceedings of European Conference on Computer Vision}, 2018.

\bibitem{carlini2017towards}
Nicholas Carlini and David Wagner.
\newblock Towards evaluating the robustness of neural networks.
\newblock In {\em 2017 ieee symposium on security and privacy (sp)}, pages
  39--57. IEEE, 2017.

\bibitem{chen2015microsoft}
Xinlei Chen, Hao Fang, Tsung-Yi Lin, Ramakrishna Vedantam, Saurabh Gupta, Piotr
  Doll{\'a}r, and C~Lawrence Zitnick.
\newblock Microsoft coco captions: Data collection and evaluation server.
\newblock {\em arXiv preprint arXiv:1504.00325}, 2015.

\bibitem{chen2020uniter}
Yen-Chun Chen, Linjie Li, Licheng Yu, Ahmed El~Kholy, Faisal Ahmed, Zhe Gan,
  Yu~Cheng, and Jingjing Liu.
\newblock Uniter: Universal image-text representation learning.
\newblock In {\em Proceedings of European Conference on Computer Vision}, pages
  104--120. Springer, 2020.

\bibitem{cisse2017parseval}
Moustapha Cisse, Piotr Bojanowski, Edouard Grave, Yann Dauphin, and Nicolas
  Usunier.
\newblock Parseval networks: Improving robustness to adversarial examples.
\newblock In {\em Proceedings on International Conference on Machine Learning},
  pages 854--863. PMLR, 2017.

\bibitem{deng2009imagenet}
Jia Deng, Wei Dong, Richard Socher, Li-Jia Li, Kai Li, and Li~Fei-Fei.
\newblock Imagenet: A large-scale hierarchical image database.
\newblock In {\em Proceedings of IEEE/CVF Conference on Computer Vision and
  Pattern Recognition}, 2009.

\bibitem{devlin2018bert}
Jacob Devlin, Ming-Wei Chang, Kenton Lee, and Kristina Toutanova.
\newblock Bert: Pre-training of deep bidirectional transformers for language
  understanding.
\newblock {\em arXiv preprint arXiv:1810.04805}, 2018.

\bibitem{dinan2019build}
Emily Dinan, Samuel Humeau, Bharath Chintagunta, and Jason Weston.
\newblock Build it break it fix it for dialogue safety: Robustness from
  adversarial human attack.
\newblock In {\em Proceedings of Empirical Methods in Natural Language
  Processing}, 2019.

\bibitem{fukui2016multimodal}
Akira Fukui, Dong~Huk Park, Daylen Yang, Anna Rohrbach, Trevor Darrell, and
  Marcus Rohrbach.
\newblock Multimodal compact bilinear pooling for visual question answering and
  visual grounding.
\newblock In {\em Proceedings of Empirical Methods in Natural Language
  Processing}, 2016.

\bibitem{gan2020large}
Zhe Gan, Yen-Chun Chen, Linjie Li, Chen Zhu, Yu~Cheng, and Jingjing Liu.
\newblock Large-scale adversarial training for vision-and-language
  representation learning.
\newblock In {\em Proceedings of Conference on Advances in Neural Information
  Processing Systems}, 2020.

\bibitem{goodfellow2014adversarial}
Ian~J Goodfellow, Jonathon Shlens, and Christian Szegedy.
\newblock Explaining and harnessing adversarial examples.
\newblock In {\em Proceedings of International Conference on Learning
  Representations}, 2015.

\bibitem{vqa2}
Yash Goyal, Tejas Khot, Douglas Summers{-}Stay, Dhruv Batra, and Devi Parikh.
\newblock Making the {V} in {VQA} matter: Elevating the role of image
  understanding in {V}isual {Q}uestion {A}nswering.
\newblock In {\em Proceedings of IEEE/CVF Conference on Computer Vision and
  Pattern Recognition}, 2017.

\bibitem{gurari2018vizwiz}
Danna Gurari, Qing Li, Abigale~J. Stangl, Anhong Guo, Chi Lin, Kristen Grauman,
  Jiebo Luo, and Jeffrey~P. Bigham.
\newblock Vizwiz grand challenge: Answering visual questions from blind people.
\newblock In {\em Proceedings of IEEE/CVF Conference on Computer Vision and
  Pattern Recognition}, 2018.

\bibitem{he2016deep}
Kaiming He, Xiangyu Zhang, Shaoqing Ren, and Jian Sun.
\newblock Deep residual learning for image recognition.
\newblock In {\em Proceedings of IEEE/CVF Conference on Computer Vision and
  Pattern Recognition}, pages 770--778, 2016.

\bibitem{hochreiter1997long}
Sepp Hochreiter and J{\"u}rgen Schmidhuber.
\newblock Long short-term memory.
\newblock {\em Neural computation}, 9(8):1735--1780, 1997.

\bibitem{hu2021unit}
Ronghang Hu and Amanpreet Singh.
\newblock Unit: Multimodal multitask learning with a unified transformer.
\newblock {\em arXiv preprint arXiv:2102.10772}, 2021.

\bibitem{hu2020iterative}
Ronghang Hu, Amanpreet Singh, Trevor Darrell, and Marcus Rohrbach.
\newblock Iterative answer prediction with pointer-augmented multimodal
  transformers for textvqa.
\newblock In {\em Proceedings of IEEE/CVF Conference on Computer Vision and
  Pattern Recognition}, pages 9992--10002, 2020.

\bibitem{gqa}
Drew~A. Hudson and Christopher~D. Manning.
\newblock Gqa: A new dataset for real-world visual reasoning and compositional
  question answering.
\newblock In {\em Proceedings of IEEE/CVF Conference on Computer Vision and
  Pattern Recognition}, 2019.

\bibitem{jiang2018pythia}
Yu~Jiang, Vivek Natarajan, Xinlei Chen, Marcus Rohrbach, Dhruv Batra, and Devi
  Parikh.
\newblock Pythia v0. 1: the winning entry to the vqa challenge 2018.
\newblock {\em arXiv preprint arXiv:1807.09956}, 2018.

\bibitem{clevr}
Justin Johnson, Bharath Hariharan, Laurens van~der Maaten, Li~Fei{-}Fei,
  C.~Lawrence Zitnick, and Ross~B. Girshick.
\newblock {CLEVR:} {A} diagnostic dataset for compositional language and
  elementary visual reasoning.
\newblock In {\em Proceedings of IEEE/CVF Conference on Computer Vision and
  Pattern Recognition}, 2017.

\bibitem{kiela2021dyna}
Douwe Kiela, Max Bartolo, Yixin Nie, Divyansh Kaushik, Atticus Geiger,
  Zhengxuan Wu, Bertie Vidgen, Grusha Prasad, Amanpreet Singh, Pratik Ringshia,
  Zhiyi Ma, Tristan Thrush, Sebastian Riedel, Zeerak Waseem, Pontus Stenetorp,
  Robin Jia, Mohit Bansal, Christopher Potts, and Adina Williams.
\newblock Dynabench: Rethinking benchmarking in nlp.
\newblock In {\em Proceedings of Annual Conference of the North American
  Chapter of the Association for Computational Linguistics}, 2021.

\bibitem{kiela2019supervised}
Douwe Kiela, Suvrat Bhooshan, Hamed Firooz, Ethan Perez, and Davide Testuggine.
\newblock Supervised multimodal bitransformers for classifying images and text.
\newblock {\em arXiv preprint arXiv:1909.02950}, 2019.

\bibitem{hateful}
Douwe Kiela, Hamed Firooz, Aravind Mohan, Vedanuj Goswami, Amanpreet Singh,
  Pratik Ringshia, and Davide Testuggine.
\newblock The hateful memes challenge: Detecting hate speech in multimodal
  memes.
\newblock In {\em Proceedings of Conference on Advances in Neural Information
  Processing Systems}, 2021.

\bibitem{kim2021vilt}
Wonjae Kim, Bokyung Son, and Ildoo Kim.
\newblock Vilt: Vision-and-language transformer without convolution or region
  supervision.
\newblock {\em arXiv preprint arXiv:2102.03334}, 2021.

\bibitem{kingma2014adam}
Diederik~P Kingma and Jimmy Ba.
\newblock Adam: A method for stochastic optimization.
\newblock In {\em Proceedings of International Conference on Learning
  Representations}, 2015.

\bibitem{krishna2017visual}
Ranjay Krishna, Yuke Zhu, Oliver Groth, Justin Johnson, Kenji Hata, Joshua
  Kravitz, Stephanie Chen, Yannis Kalantidis, Li-Jia Li, David~A Shamma, et~al.
\newblock Visual genome: Connecting language and vision using crowdsourced
  dense image annotations.
\newblock {\em Proceedings of International Journal of Computer Vision},
  123(1):32--73, 2017.

\bibitem{li2019visualbert}
Liunian~Harold Li, Mark Yatskar, Da~Yin, Cho-Jui Hsieh, and Kai-Wei Chang.
\newblock Visualbert: A simple and performant baseline for vision and language.
\newblock In {\em Proceedings of Annual Meeting of the Association for
  Computational Linguistics}, 2020.

\bibitem{li2020unimo}
Wei Li, Can Gao, Guocheng Niu, Xinyan Xiao, Hao Liu, Jiachen Liu, Hua Wu, and
  Haifeng Wang.
\newblock Unimo: Towards unified-modal understanding and generation via
  cross-modal contrastive learning.
\newblock {\em arXiv preprint arXiv:2012.15409}, 2020.

\bibitem{li2020oscar}
Xiujun Li, Xi~Yin, Chunyuan Li, Pengchuan Zhang, Xiaowei Hu, Lei Zhang, Lijuan
  Wang, Houdong Hu, Li~Dong, Furu Wei, et~al.
\newblock Oscar: Object-semantics aligned pre-training for vision-language
  tasks.
\newblock In {\em Proceedings of European Conference on Computer Vision}, pages
  121--137. Springer, 2020.

\bibitem{lin2014coco}
Tsung-Yi Lin, Michael Maire, Serge Belongie, James Hays, Pietro Perona, Deva
  Ramanan, Piotr Doll{\'a}r, and C~Lawrence Zitnick.
\newblock Microsoft coco: Common objects in context.
\newblock In {\em Proceedings of European Conference on Computer Vision}, pages
  740--755. Springer, 2014.

\bibitem{lin2014microsoft}
Tsung-Yi Lin, Michael Maire, Serge Belongie, James Hays, Pietro Perona, Deva
  Ramanan, Piotr Doll{\'a}r, and C~Lawrence Zitnick.
\newblock Microsoft coco: Common objects in context.
\newblock In {\em European conference on computer vision}, pages 740--755.
  Springer, 2014.

\bibitem{loshchilov2017decoupled}
Ilya Loshchilov and Frank Hutter.
\newblock Decoupled weight decay regularization.
\newblock In {\em Proceedings of International Conference on Learning
  Representations}, 2019.

\bibitem{lu2019vilbert}
Jiasen Lu, Dhruv Batra, Devi Parikh, and Stefan Lee.
\newblock Vilbert: Pretraining task-agnostic visiolinguistic representations
  for vision-and-language tasks.
\newblock {\em arXiv preprint arXiv:1908.02265}, 2019.

\bibitem{lu202012}
Jiasen Lu, Vedanuj Goswami, Marcus Rohrbach, Devi Parikh, and Stefan Lee.
\newblock 12-in-1: Multi-task vision and language representation learning.
\newblock In {\em Proceedings of IEEE/CVF Conference on Computer Vision and
  Pattern Recognition}, pages 10437--10446, 2020.

\bibitem{okvqa}
Kenneth Marino, Mohammad Rastegari, Ali Farhadi, and Roozbeh Mottaghi.
\newblock Ok-vqa: A visual question answering benchmark requiring external
  knowledge.
\newblock In {\em Proceedings of IEEE/CVF Conference on Computer Vision and
  Pattern Recognition}, 2019.

\bibitem{nguyen2020revisiting}
Duy-Kien Nguyen, Vedanuj Goswami, and Xinlei Chen.
\newblock Revisiting modulated convolutions for visual counting and beyond.
\newblock In {\em Proceedings of International Conference on Learning
  Representations}, 2021.

\bibitem{nie2020anli}
Yixin Nie, Adina Williams, Emily Dinan, Mohit Bansal, Jason Weston, and Douwe
  Kiela.
\newblock Adversarial nli: A new benchmark for natural language understanding.
\newblock In {\em Proceedings of Annual Meeting of the Association for
  Computational Linguistics}, 2020.

\bibitem{ordonez2011im2text}
Vicente Ordonez, Girish Kulkarni, and Tamara Berg.
\newblock Im2text: Describing images using 1 million captioned photographs.
\newblock {\em Proceedings of Conference on Advances in Neural Information
  Processing Systems}, 24:1143--1151, 2011.

\bibitem{paszke2019pytorch}
Adam Paszke, Sam Gross, Francisco Massa, Adam Lerer, James Bradbury, Gregory
  Chanan, Trevor Killeen, Zeming Lin, Natalia Gimelshein, Luca Antiga, Alban
  Desmaison, Andreas K{\"{o}}pf, Edward Yang, Zach DeVito, Martin Raison,
  Alykhan Tejani, Sasank Chilamkurthy, Benoit Steiner, Lu~Fang, Junjie Bai, and
  Soumith Chintala.
\newblock Pytorch: An imperative style, high-performance deep learning library.
\newblock In {\em Proceedings of Conference on Advances in Neural Information
  Processing Systems}, 2019.

\bibitem{potts2020dynasent}
Christopher Potts, Zhengxuan Wu, Atticus Geiger, and Douwe Kiela.
\newblock Dynasent: A dynamic benchmark for sentiment analysis.
\newblock In {\em Proceedings of Annual Meeting of the Association for
  Computational Linguistics}, 2021.

\bibitem{rodriguez2019quizbowl}
Pedro Rodriguez, Shi Feng, Mohit Iyyer, He~He, and Jordan Boyd-Graber.
\newblock Quizbowl: The case for incremental question answering.
\newblock {\em arXiv preprint arXiv:1904.04792}, 2019.

\bibitem{sharma2018conceptual}
Piyush Sharma, Nan Ding, Sebastian Goodman, and Radu Soricut.
\newblock Conceptual captions: A cleaned, hypernymed, image alt-text dataset
  for automatic image captioning.
\newblock In {\em Proceedings of Annual Meeting of the Association for
  Computational Linguistics}, pages 2556--2565, 2018.

\bibitem{simonyan2014very}
Karen Simonyan and Andrew Zisserman.
\newblock Very deep convolutional networks for large-scale image recognition.
\newblock {\em arXiv preprint arXiv:1409.1556}, 2014.

\bibitem{singh2020mmf}
Amanpreet Singh, Vedanuj Goswami, Vivek Natarajan, Yu~Jiang, Xinlei Chen, Meet
  Shah, Marcus Rohrbach, Dhruv Batra, and Devi Parikh.
\newblock Mmf: A multimodal framework for vision and language research.
\newblock \url{https://github.com/facebookresearch/mmf}, 2020.

\bibitem{singh2020we}
Amanpreet Singh, Vedanuj Goswami, and Devi Parikh.
\newblock Are we pretraining it right? digging deeper into visio-linguistic
  pretraining.
\newblock {\em arXiv preprint arXiv:2004.08744}, 2020.

\bibitem{textvqa}
Amanpreet Singh, Vivek Natarajan, Meet Shah, Yu~Jiang, Xinlei Chen, Dhruv
  Batra, Devi Parikh, and Marcus Rohrbach.
\newblock Towards vqa models that can read.
\newblock In {\em Proceedings of IEEE/CVF Conference on Computer Vision and
  Pattern Recognition}, 2019.

\bibitem{trott2017interpretable}
Alexander Trott, Caiming Xiong, and Richard Socher.
\newblock Interpretable counting for visual question answering.
\newblock In {\em Proceedings of International Conference on Learning
  Representations}, 2018.

\bibitem{vaswani2017attention}
Ashish Vaswani, Noam Shazeer, Niki Parmar, Jakob Uszkoreit, Llion Jones,
  Aidan~N Gomez, Lukasz Kaiser, and Illia Polosukhin.
\newblock Attention is all you need.
\newblock In {\em Proceedings of Conference on Advances in Neural Information
  Processing Systems}, 2017.

\bibitem{vidgen2020learning}
Bertie Vidgen, Tristan Thrush, Zeerak Waseem, and Douwe Kiela.
\newblock Learning from the worst: Dynamically generated datasets to improve
  online hate detection.
\newblock In {\em Proceedings of Annual Meeting of the Association for
  Computational Linguistics}, 2021.

\bibitem{yu2019deep}
Zhou Yu, Jun Yu, Yuhao Cui, Dacheng Tao, and Qi~Tian.
\newblock Deep modular co-attention networks for visual question answering.
\newblock In {\em Proceedings of the IEEE/CVF Conference on Computer Vision and
  Pattern Recognition}, pages 6281--6290, 2019.

\bibitem{zhang2021vinvl}
Pengchuan Zhang, Xiujun Li, Xiaowei Hu, Jianwei Yang, Lei Zhang, Lijuan Wang,
  Yejin Choi, and Jianfeng Gao.
\newblock Vinvl: Making visual representations matter in vision-language
  models.
\newblock In {\em Proceedings of IEEE/CVF Conference on Computer Vision and
  Pattern Recognition}, 2021.

\bibitem{biasemnlp}
Jieyu Zhao, Tianlu Wang, Mark Yatskar, Vicente Ordonez, and Kai-Wei Chang.
\newblock Men also like shopping: Reducing gender bias amplification using
  corpus-level constraints.
\newblock In {\em Proceedings of Empirical Methods in Natural Language
  Processing}, 2017.

\end{thebibliography}
\newpage
\appendix
\counterwithin{figure}{section}
\counterwithin{table}{section}
\counterwithin{equation}{section}
{
\begin{center}
\Large 
\textbf{Human-Adversarial Visual Question Answering}
\\
(Supplementary Material)
\par
\end{center}
\vspace{2em}
}

\section{Training Details}
Except UNITER \cite{chen2020uniter} and VILLA \cite{gan2020large}, we trained all models using MMF \cite{singh2020mmf}.
The evaluation results with the standard deviation over three different runs with three different seeds are provided in Table \ref{tab:result_b}. Below we detail the finetuning setup on the COCO \cite{lin2014coco} train 2017 split evaluated on the AdVQA validation split. Unless otherwise specified, we used the AdamW \cite{kingma2014adam,loshchilov2017decoupled} optimizer with an initial learning rate of $5e-5$, epsilon of $1e-8$, cosine schedule and a warmup of 2000 steps. All jobs are trained in distributed fashion using PyTorch \cite{paszke2019pytorch}› on NVIDIA V100 GPUs.

\begin{table}[h]
    \centering
    \footnotesize
    \setlength{\tabcolsep}{2.9pt}
    \caption{\textbf{Model performance on VQA v2 and \acronym{}} val and test sets. $^*$ indicates that this model architecture (but not this model instance) was used in the data collection loop.}
    \begin{tabular}{@{}cl|rr|rr@{}}
        \toprule
         \multicolumn{2}{c|}{\textbf{Model}} & \multicolumn{1}{c}{\textbf{VQA}} & \multicolumn{1}{c|}{\textbf{\acronym{}}} & \multicolumn{1}{c}{\textbf{VQA}} & \multicolumn{1}{c}{\textbf{\acronym{}}} \\
        & & \multicolumn{1}{c}{test-dev} & \multicolumn{1}{c|}{test} & \multicolumn{2}{c}{val} \\\midrule
        \multicolumn{2}{c|}{\emph{Human performance}} & 80.78 & 91.18 & 84.73 & 87.53 \\\midrule
        \multicolumn{2}{c|}{\emph{Majority answer (overall)}} & - & 13.38 & 24.67 & 11.65 \\
        \multicolumn{2}{c|}{\emph{Majority answer (per answer type)}} & - & 27.39 & 31.01 & 29.24 \\\midrule
        Model in loop & MoViE+MCAN \cite{nguyen2020revisiting} & 73.56 & 10.33  & 73.51  & 10.24  \\
        \midrule
        \multirow{2}{*}{Unimodal} & ResNet-152 \cite{he2016deep} & 26.37±0.38 & 10.85±0.37 & 24.82±0.27 & 11.22±0.23 \\
        & BERT \cite{devlin2018bert} &  39.47±2.92 & 26.90±0.36 & 39.40±3.23 & 23.81±0.86 \\\midrule
        \multirow{3}{*}{\shortstack{Multimodal\\(unimodal pretrain)}} & MoViE+MCAN$^*$ \cite{nguyen2020revisiting} & 71.36±0.27  & 26.64±0.45  & 71.31±0.13  & 26.37±0.49  \\
        & MMBT \cite{kiela2019supervised} & 58.00±4.10  & 26.70±0.24 & 57.32±3.75 & 25.78±0.34 \\
        & UniT \cite{hu2021unit} & 64.36±0.13 & 28.15±0.21  & 64.32±0.08  & 27.55±0.16  \\
        \midrule
        \multirow{6}{*}{\shortstack{Multimodal\\(multimodal pretrain)}} & VisualBERT \cite{li2019visualbert} & 70.37±0.05  & 28.70±0.36  & 70.05±0.11  & 28.03±0.33  \\
        & ViLBERT \cite{lu2019vilbert} & 69.42±0.30  & 27.36±0.18 & 69.27±0.14  & 27.36±0.32 \\
        & ViLT \cite{kim2021vilt} & 64.52±0.42  & 27.11±0.14  & 65.43±2.43  & 27.19±0.50  \\
        & UNITER$_{Base}$ \cite{chen2020uniter} & 71.87±0.01 & 25.16±0.49 & 70.50±0.08 & 25.20±0.19 \\
        & UNITER$_{Large}$ \cite{chen2020uniter} & 73.57±0.21 & 26.94±0.31 & 72.71±0.22 & 28.03±0.33 \\
        & VILLA$_{Base}$ \cite{gan2020large} & 70.94±1.25 & 25.14±0.93 & 69.50±1.44 & 25.17±0.67 \\
        & VILLA$_{Large}$ \cite{gan2020large} & 72.29±0.39 & 25.79±0.46 & 71.40±0.37 &  26.18±0.15 \\
        \midrule
        \multirow{2}{*}{\shortstack{Multimodal\\(unimodal pretrain + OCR)}} & M4C (TextVQA+STVQA) \cite{hu2020iterative} & 32.89±0.57  & 28.86±0.35  & 31.44±0.59  & 29.08±0.33  \\
        & M4C (VQA v2 train set) \cite{hu2020iterative} & 67.66±0.34  & 33.52±0.47  & 66.21±0.38  & 33.33±0.51 \\
        \bottomrule 
    \end{tabular}
    \label{tab:result_b}
\end{table}

\textbf{MoViE+MCAN} We use a batch size of 64 for 236K updates using a multi-step learning rate scheduler with steps at 180K and 216K, learning rate ratio of 0.2 and a warmup for 54K updates. Training takes an average of 2 days.

\textbf{Unimodal} We train the models with a batch size of 64 for 88K updates with linear learning rate schedule starting from $1e-5$ with a warmup for 2000 updates. We used a linear learning rate schedule with 2000 warm up steps. The training takes an average of 8 hours.

\textbf{MMBT~\cite{kiela2019supervised}} We trained MMBT from scratch with a batch size 64 without any pretraining following \cite{kiela2019supervised} for 88K updates. The training takes an average of 17 hours.

\textbf{UniT~\cite{hu2021unit}} We initialized from the model pretrained on all 8 datasets \cite{hu2021unit} with COCO initialization. We set the batch size to 8, weight decay as $1e-4$ and train the model on 8 GPUs for 2 days.

\textbf{VisualBERT~\cite{li2019visualbert}} We trained VisualBERT from the best pretrained model on COCO using MLM loss using a batch size of 64 which takes an average of 8 hours.

\textbf{ViLBERT~\cite{lu2019vilbert}} We trained ViLBERT from the best pretrained model on Conceptual Captions using MLM loss using a batch size of 64 which takes an average of 13 hours.

\textbf{ViLT~\cite{kim2021vilt}} We trained ViLT with 44K updates, initial learning rate of $1e-4$, eps as $1e-8$ and weight decay as 0.01 for an average of 7 hours.

\textbf{UNITER/VILLA~\cite{chen2020uniter, gan2020large}} We used the author-provided pretrained checkpoint for UNITER\footnote{UNITER Code: https://github.com/ChenRocks/UNITER} and VILLA\footnote{VILLA Code: https://github.com/zhegan27/VILLA} with a confidence threshold of 0.075. The rest of the hyper parameters were consistent with the configuration provided in the repository. The training takes about 2 hours.

\textbf{M4C~\cite{hu2020iterative}} We used the same training schedule and hyper parameters as \cite{hu2020iterative} used for TextVQA training \cite{textvqa} for both TextVQA + STVQA and VQA v2. We didn't use extra Visual Genome \cite{krishna2017visual} QA data due to lack of OCR text and used the OCR data from \cite{hu2020iterative} for both COCO and TextVQA.

\section{Off-the-shelf models}

In addition to the main results in Table \ref{tab:result_a}, we also evaluated existing OSCAR \cite{li2020oscar} / VinVL \cite{zhang2021vinvl} models on \acronym{}, since both models are known to perform well on VQA. Note that the training set of the off-the-shelf OSCAR and VinVL models includes COCO val2014 data, which overlaps with our validation set (COCO 2017). Even then, we find that the validation performance is still much lower (more than 2x) compared to its performance on VQA v2. See Table \ref{tab:off_shelf_oscar}.

\begin{table}[h]
    \begin{center}
        \caption{\textbf{Validation performance of off-the-shelf OSCAR and VinVL models on \acronym{}}. Note that \acronym{} validation set has some overlap with the off-the-shelf models' training set.}
        \begin{tabular}{l|ll}
            \toprule
            Model & VQA & \acronym{} \\\midrule
            OSCAR$_{Base}$ & 72.59 & 29.17 \\
            OSCAR$_{Large}$ & 92.64 & 31.60 \\
            VinVL$_{Base}$ & 75.13 & 33.59 \\
            VinVL$_{Large}$ & 76.23 & 37.12 \\
            \bottomrule
        \end{tabular}
        \label{tab:off_shelf_oscar}
    \end{center}
\end{table}

\begin{figure*}[h]
    \centering
    \begin{subfigure}[b]{0.9\linewidth}
        \includegraphics[width=1\linewidth]{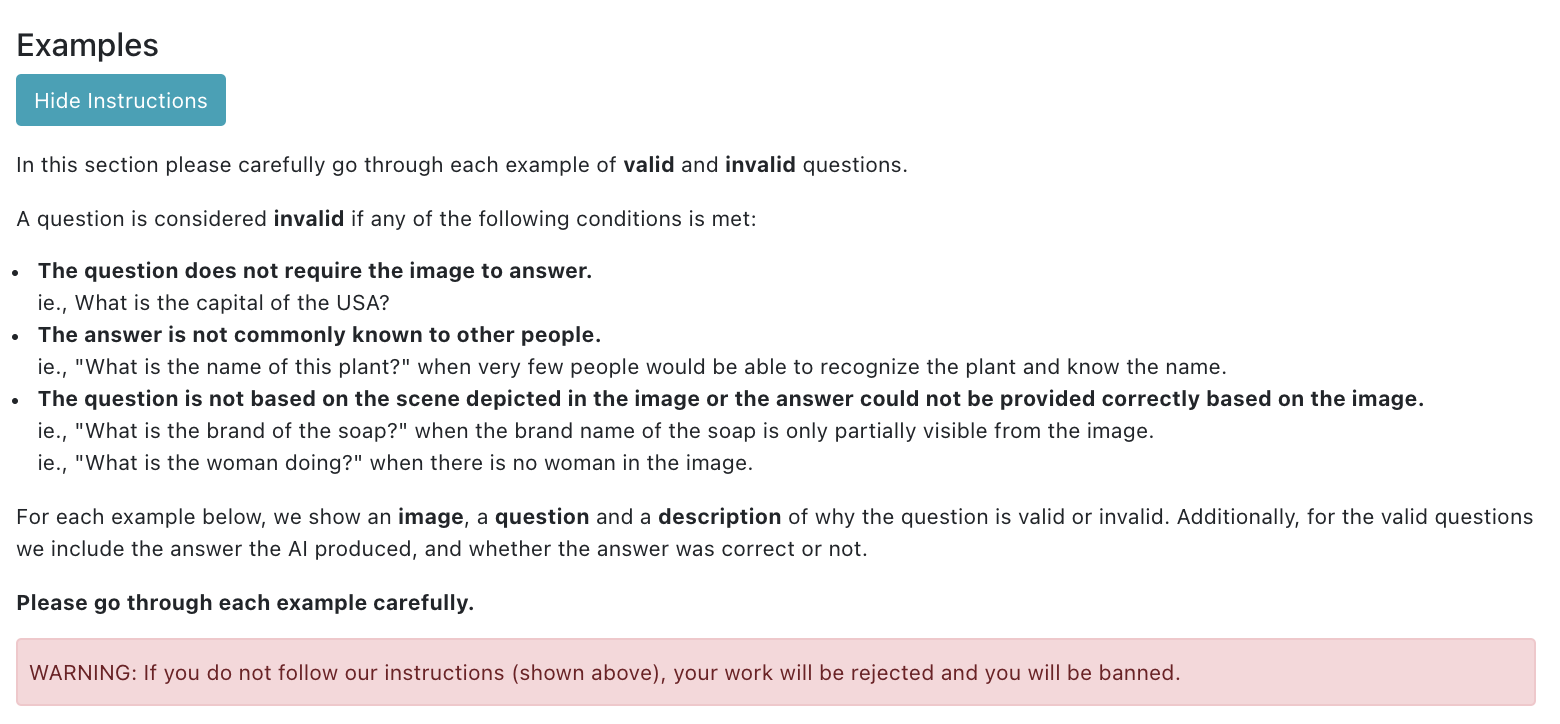}
        \caption{Instructions for Examples Stage}
        \label{fig:examples_instr}
    \end{subfigure}
    \begin{subfigure}[b]{0.45\linewidth}
        \includegraphics[width=1\linewidth]{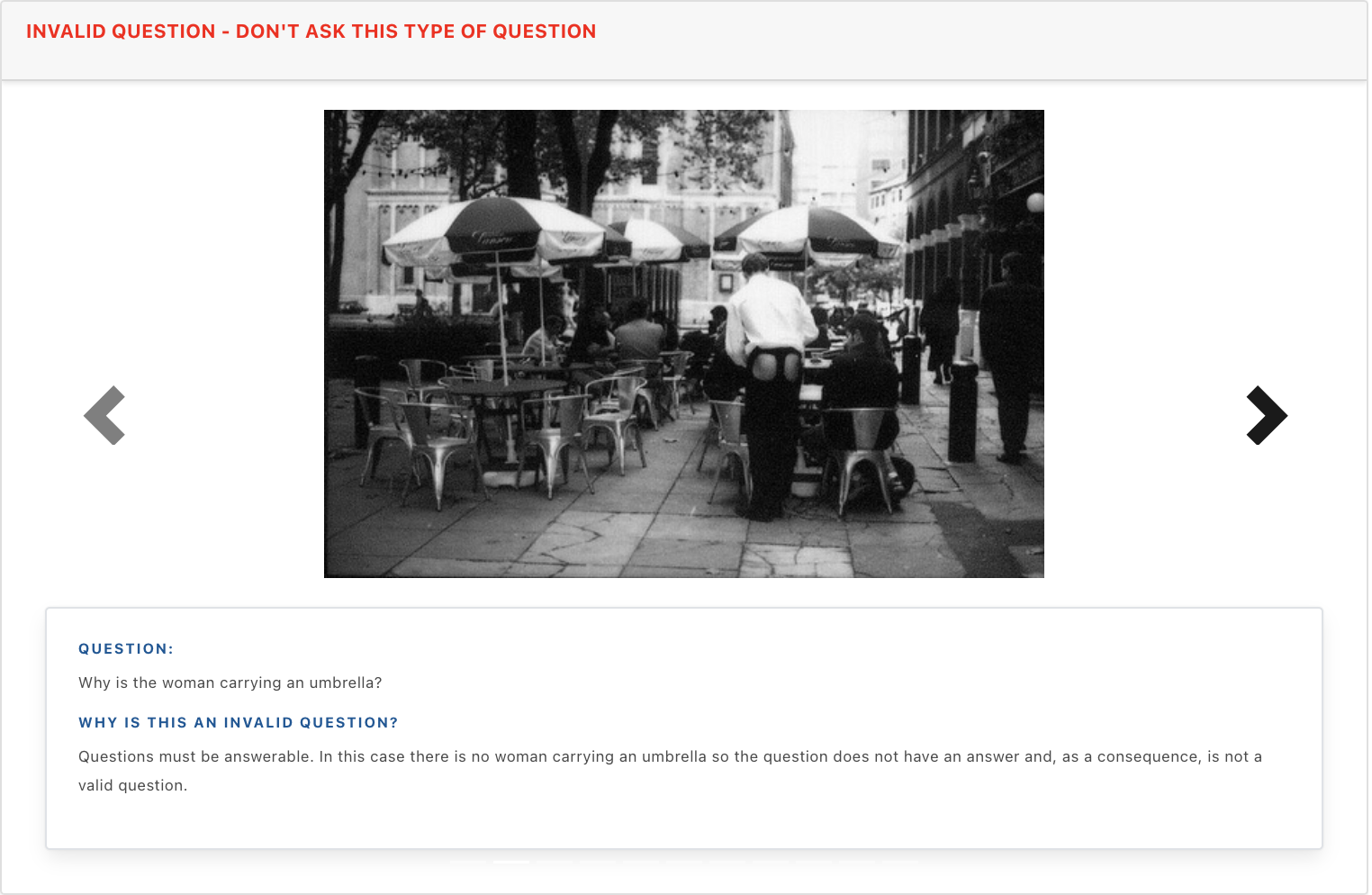}
        \caption{Invalid example}
        \label{fig:example_1}
    \end{subfigure}
    \begin{subfigure}[b]{0.45\linewidth}
        \includegraphics[width=1\linewidth]{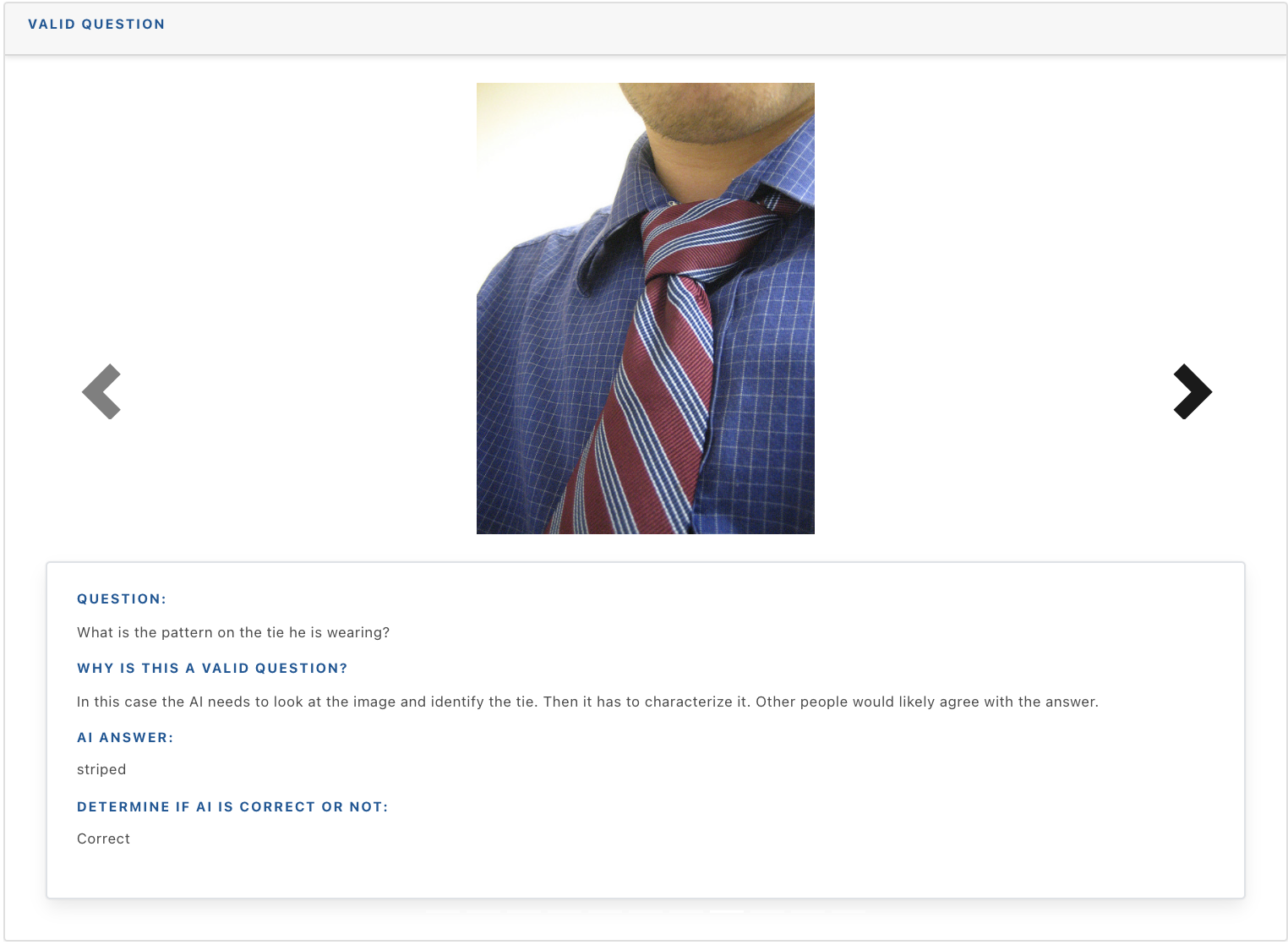}
        \caption{Valid example}
        \label{fig:fig:example_2}
    \end{subfigure}
    \caption{Qualification Phase Stage 1: View Examples}
    \label{fig:examples}
\end{figure*}

\begin{figure*}[h]
    \centering
    \begin{subfigure}[b]{0.9\linewidth}
        \includegraphics[width=1\linewidth]{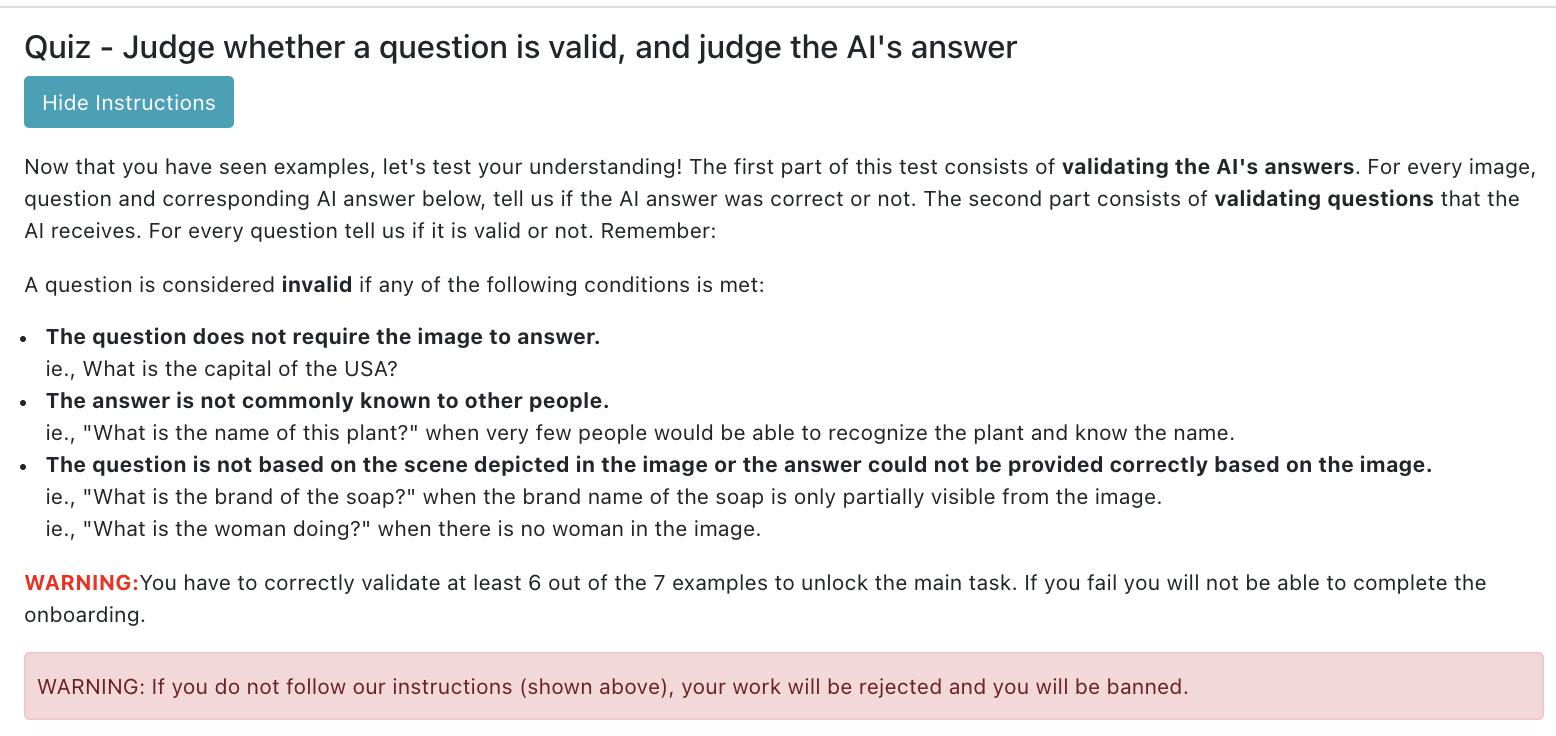}
        \caption{Instructions for Quiz Stage}
        \label{fig:quiz_instr}
    \end{subfigure}
    \begin{subfigure}[b]{0.45\linewidth}
        \includegraphics[width=1\linewidth]{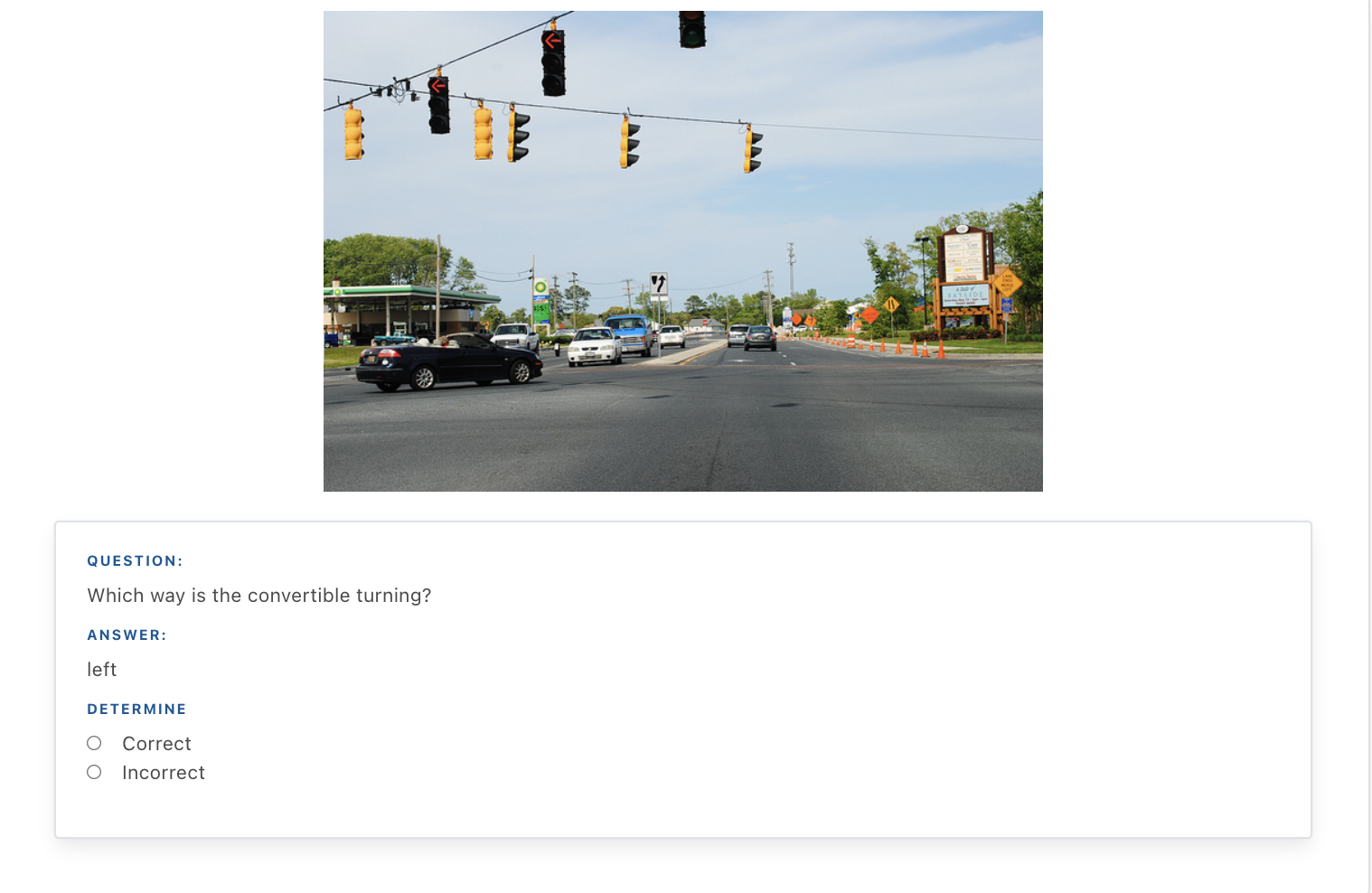}
        \caption{Determine if the answer is correct}
        \label{fig:quiz_1}
    \end{subfigure}
    \begin{subfigure}[b]{0.45\linewidth}
        \includegraphics[width=1\linewidth]{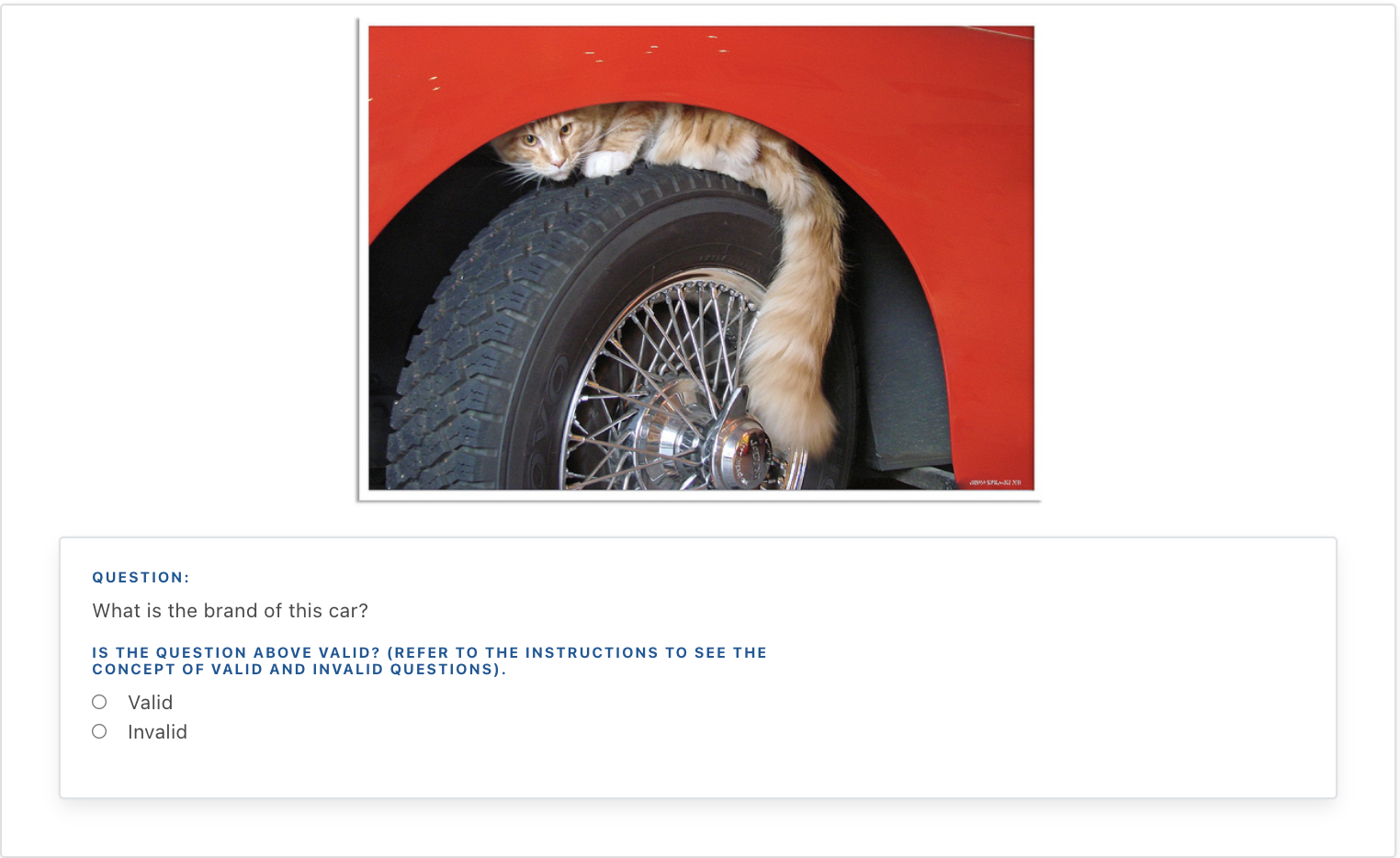}
        \caption{Determine if the question is valid}
        \label{fig:quiz_2}
    \end{subfigure}
    \caption{Qualification Phase Stage 2 - quiz. Annotators are allowed access to the main task only if they passed the quiz. }
    \label{fig:quizzes}
\end{figure*}
\begin{figure}[h]
    \begin{subfigure}[b]{0.45\linewidth}
        \includegraphics[width=1\linewidth]{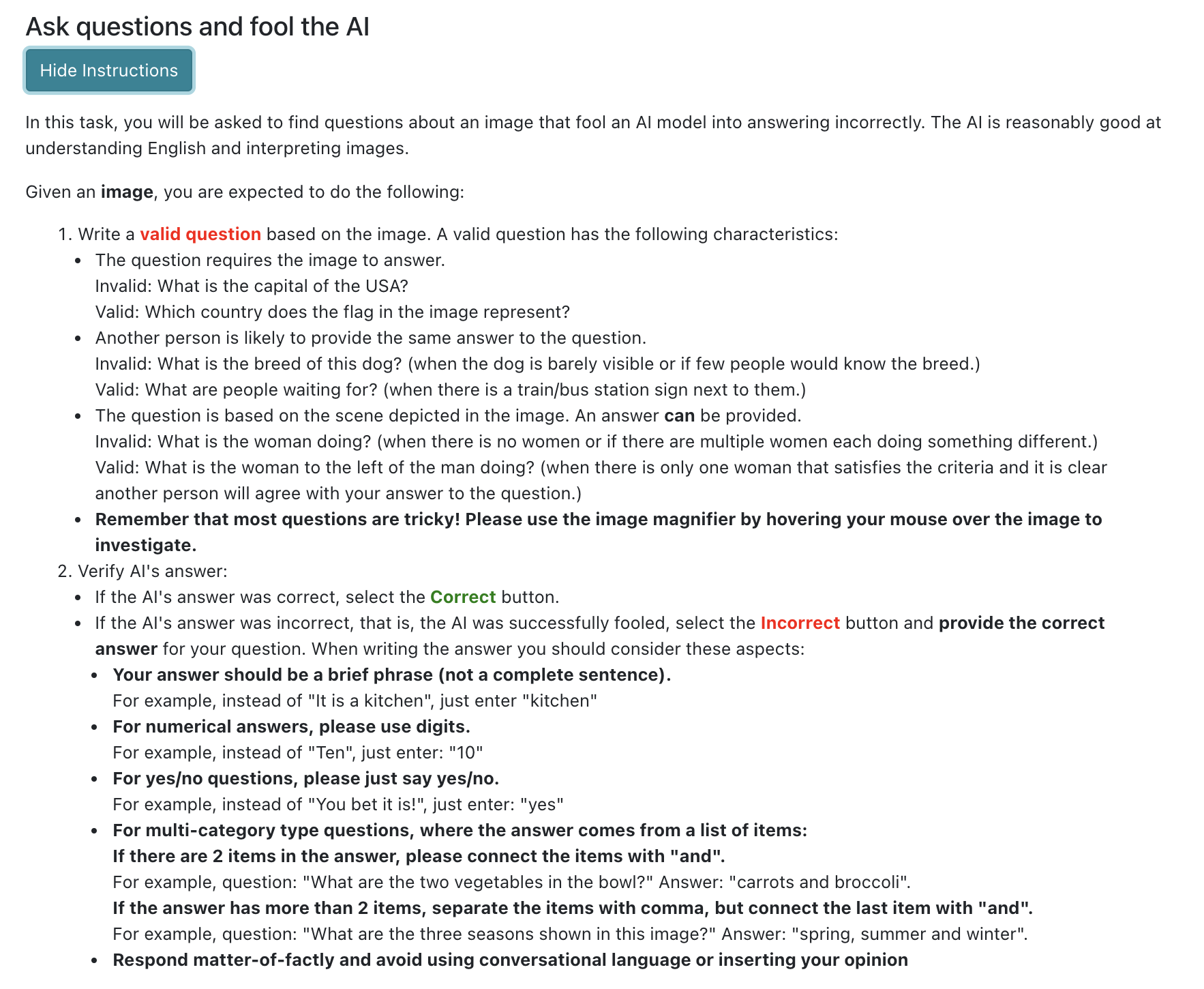}
        \caption{Question Collection Instructions}
        \label{fig:qc_instr}
    \end{subfigure}
    \begin{subfigure}[b]{0.45\linewidth}
        \includegraphics[width=1\linewidth]{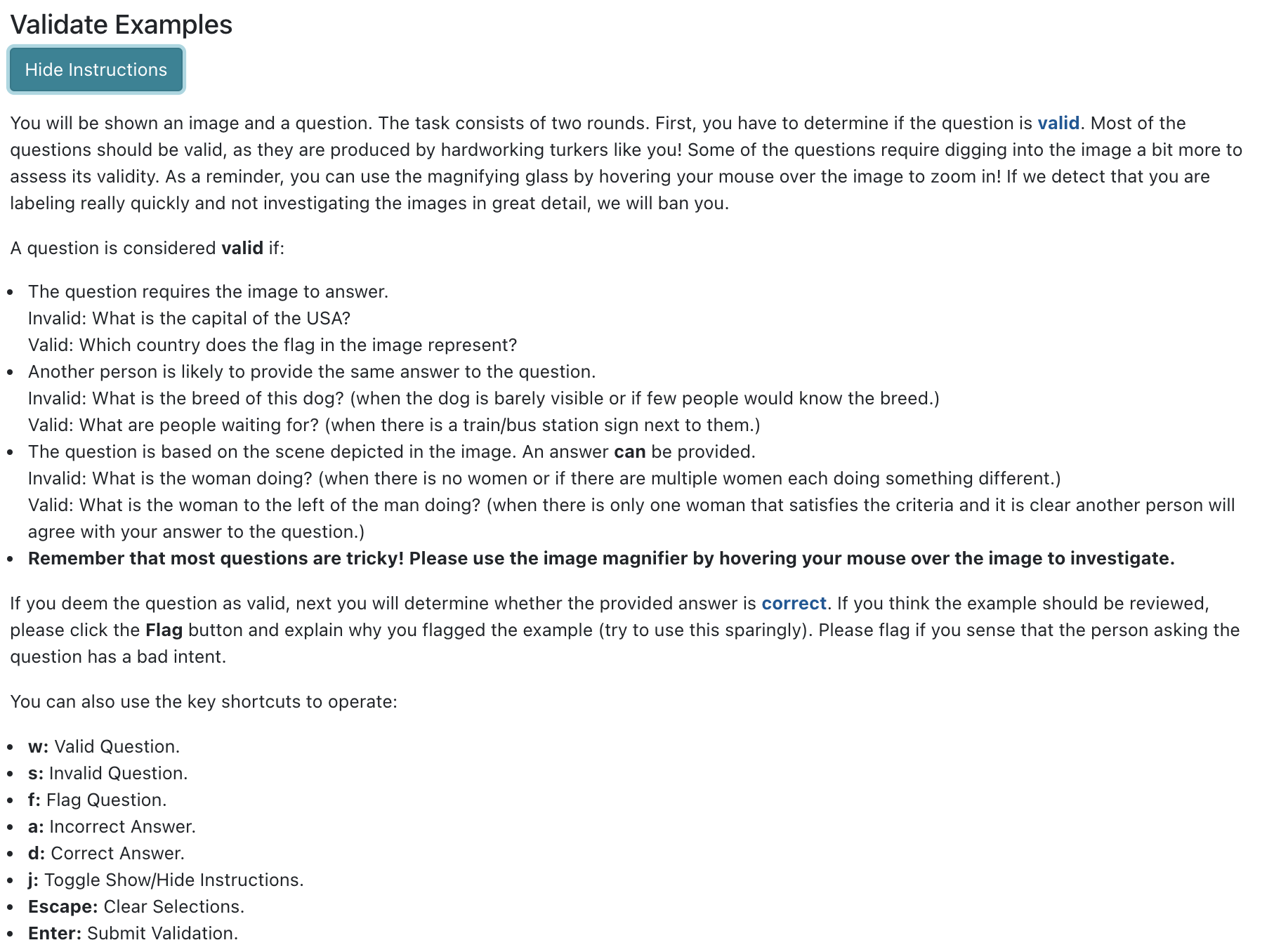}
        \caption{Question Validation Instructions}
        \label{fig:val_instr}
    \end{subfigure}
    \caption{Main Labeling Task Instructions}
\end{figure}

\begin{figure}[h]
    \begin{subfigure}[b]{0.45\linewidth}
        \includegraphics[width=1\linewidth]{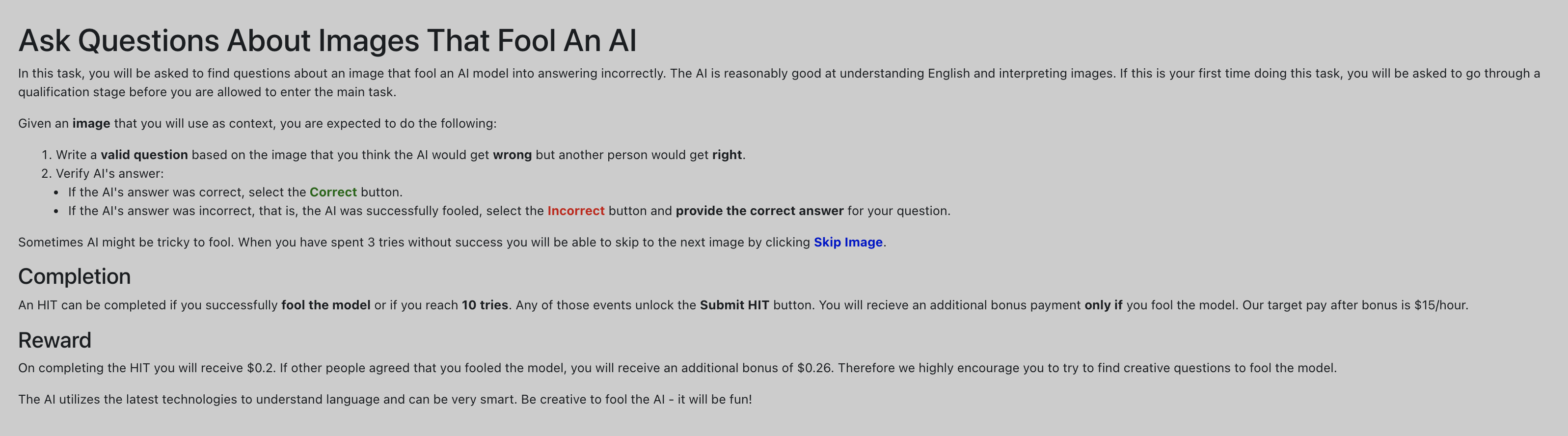}
        \caption{Preview for Question Collection}
        \label{fig:prev_qc}
    \end{subfigure}
    \begin{subfigure}[b]{0.45\linewidth}
        \includegraphics[width=1\linewidth]{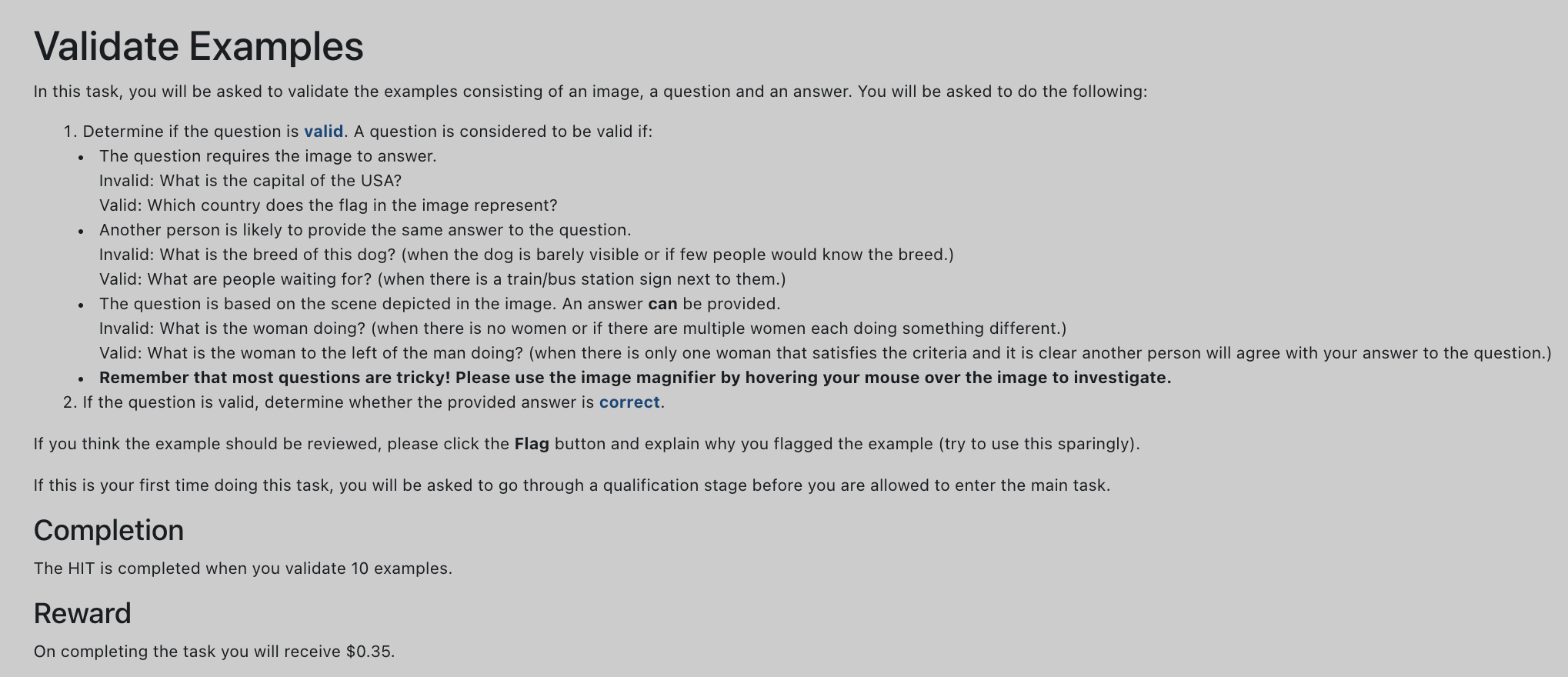}
        \caption{Preview for Question Validations}
        \label{fig:prev_val}
    \end{subfigure}
    \caption{Preview - landing page on MTurk interface.}
\end{figure}

\begin{figure}[h]
    \centering
    \includegraphics[width=1\linewidth]{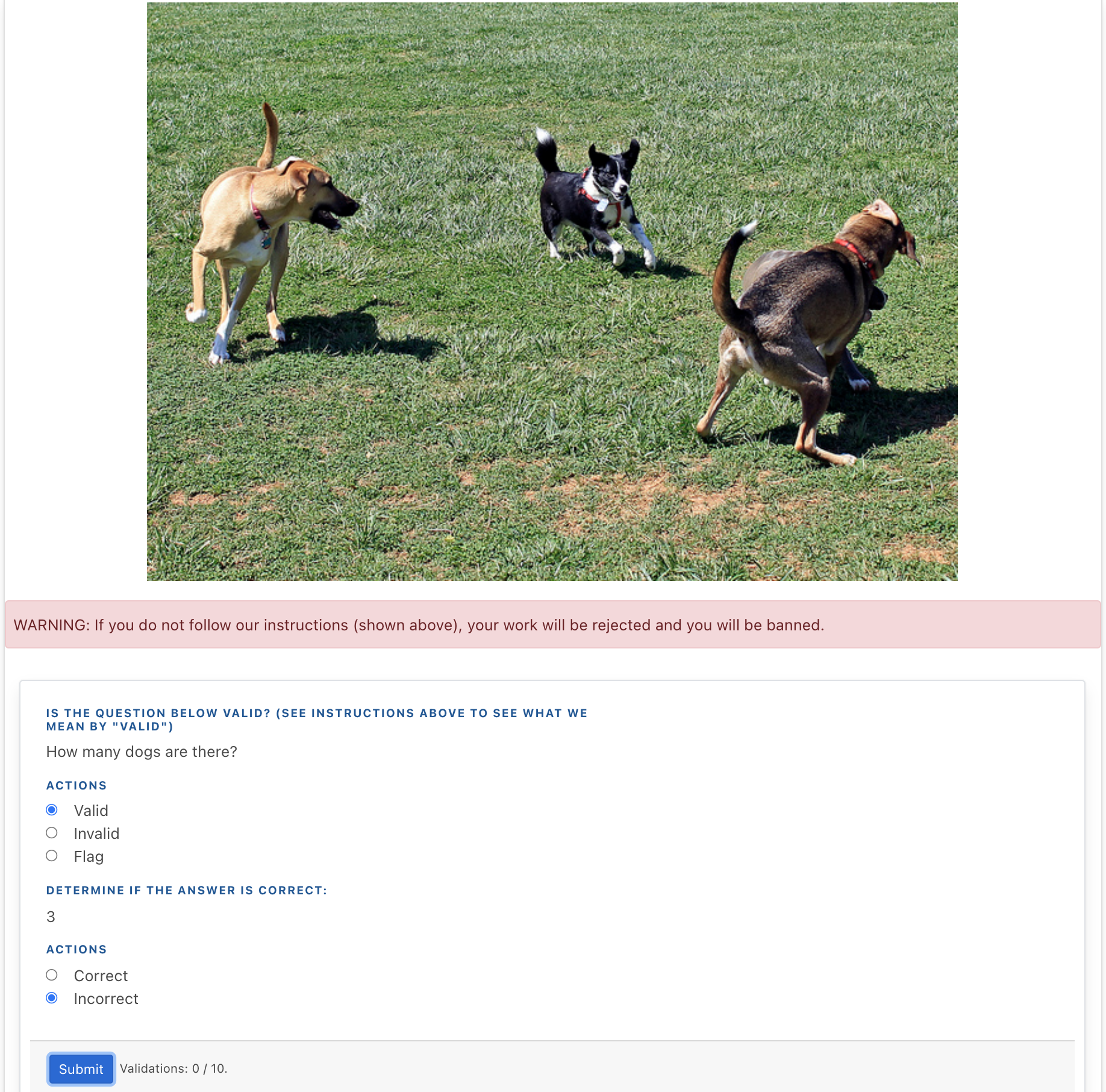}
    \caption{Validation Interface}
    \label{fig:val_interface}
\end{figure}
\section{Annotation details}
\label{appendix:annotation}
Before annotators were able to proceed with the main task, they had to pass an on-boarding phase (Section~\ref{appendix:qualification}). We restricted the interface to be only accessible on non-mobile devices, to annotators in the US with at least 100 approved hits and with an approval rate higher than 98\%. The annotators were paid a bonus if their self-claimed fooling question was verified to be fooling and valid by two other annotators.

\subsection{Qualification Phase}
\label{appendix:qualification}

The annotators were asked to go through a two-stage qualification phase. In the first stage, they were shown 11 examples that include both valid and invalid questions. Figure~\ref{fig:examples} shows valid and invalid examples in the Example Stage. After scrolling through those 11 examples, the annotators proceeded to the second stage, in which we ask them to complete a quiz. The annotators passed only if they got more than 6 out of 7 correct, after which they qualified for the main task. There are two types of quiz questions: 1) determine if the provided answer is correct for the specific image question pair; and 2) determine if a given question is valid with the image as context. See Figure~\ref{fig:quizzes} for examples of those two question types. If an annotator failed the first time, they were given an explanation on the correct choice before being allowed a second try on a different (but similar) set of questions.   

\subsection{Main Labeling Task}

Figure \ref{fig:qc_instr} and \ref{fig:val_instr} show the instructions given to first-time annotators for the ``question collection'' and ``question validation'' tasks. The instructions were hidden for the non-first-time annotators, but remained accessible via a button at the top. Figure ~\ref{fig:prev_qc} and \ref{fig:prev_val} show the preview landing pages on Mechanical Turk.

\begin{figure}[h]
    \includegraphics[width=1\linewidth]{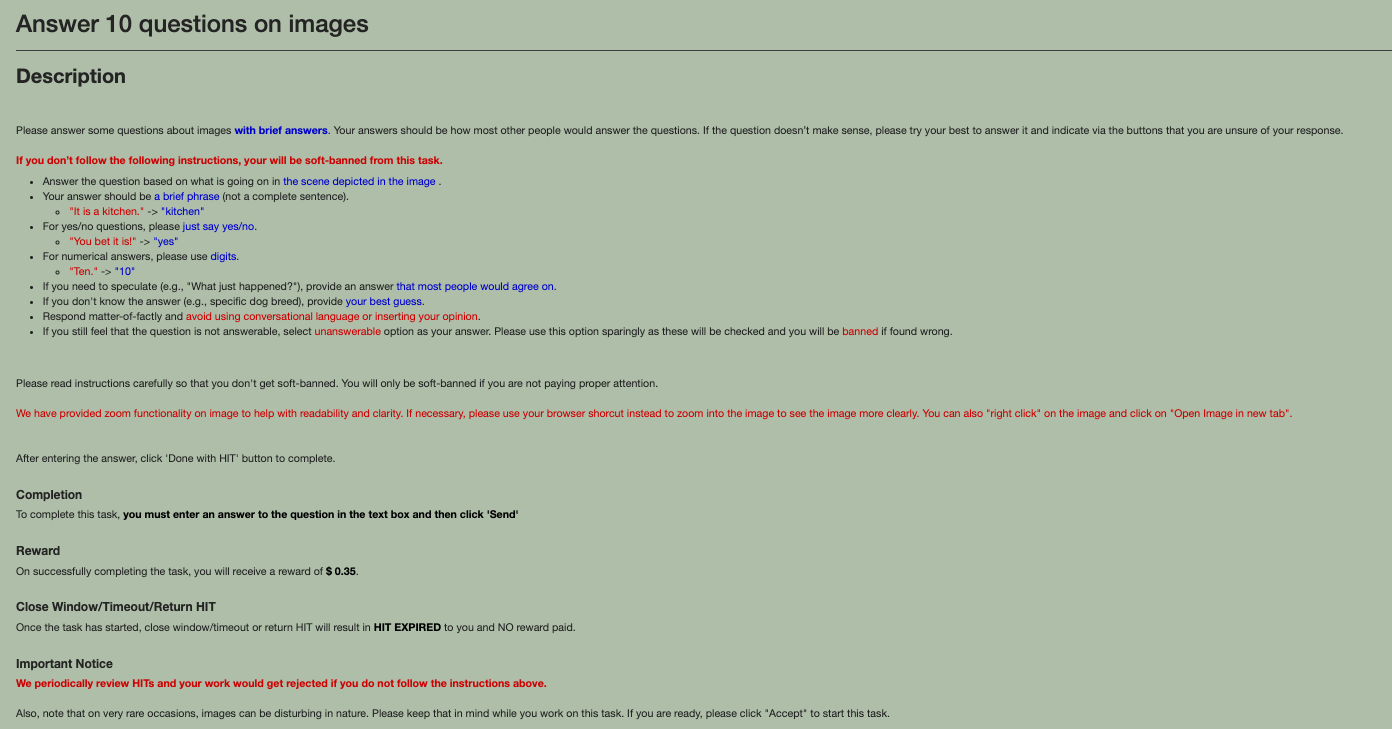}
    \caption{Preview for Answer Collection}
    \label{fig:answer_preview}
\end{figure}
\begin{figure}[h]
    \centering
    \includegraphics[width=1\linewidth]{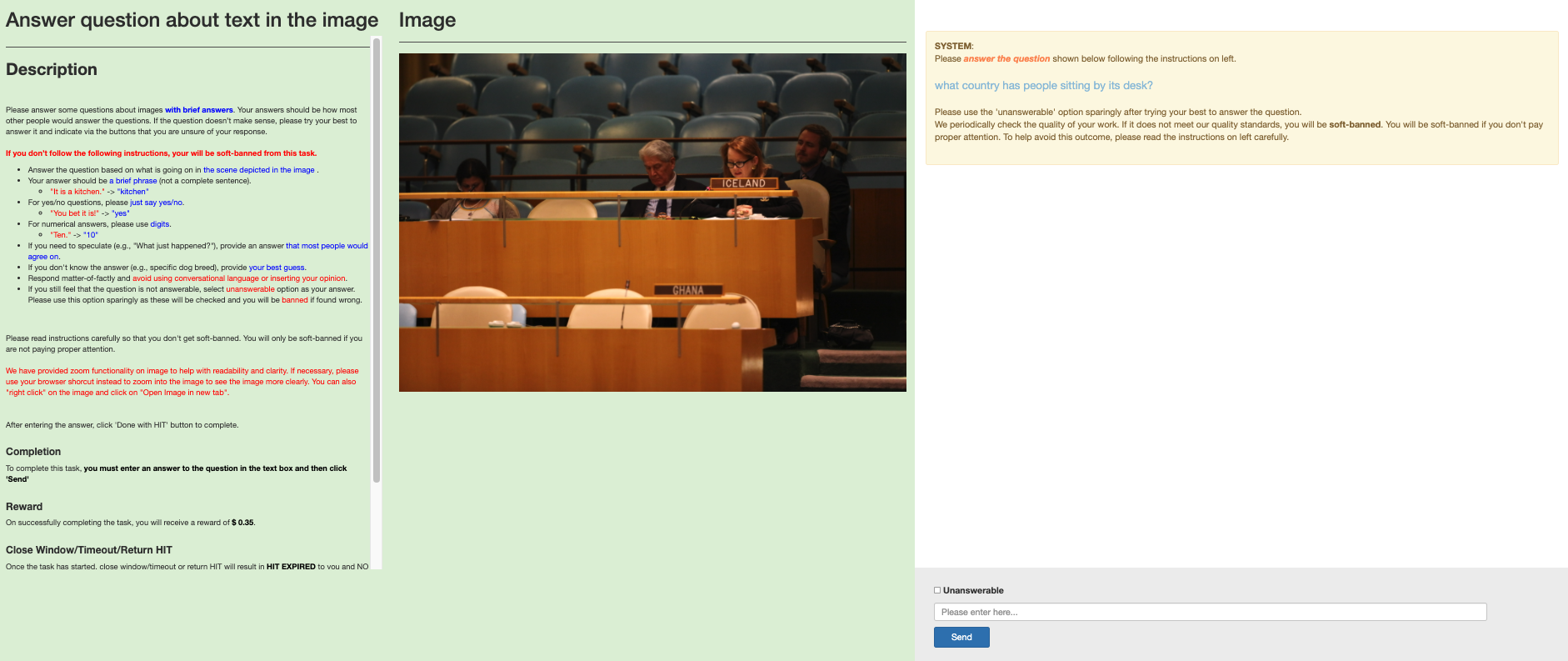}
    \caption{Answer collection interface}
    \label{fig:answer_main}
\end{figure}

\subsection{Answer collection task}

Figure~\ref{fig:answer_preview} and \ref{fig:answer_main} show the preview and main interface of the answer collection stage. For each question, we collect ten answers from ten different annotators using this interface. We provide explicit instructions following \cite{vqa2} and \cite{textvqa} to avoid ambiguity and collect short relevant answers. The annotators are also provided a checkbox to select ``unanswerable'' in case the question is ambiguous or can't be answered which annotators are suggested to use sparingly. Finally, we use and show a set of hand-crafted already annotated questions without ambiguity randomly to filter out bad annotators by comparing their answers to ground truth. An annotator is prevented from doing the task if they fail the test three times. 
\clearpage
\section{Random Samples}

We show 10 randomly selected samples in Figure \ref{tab:random_3}.

\begin{table*}[h]
    \centering
    \footnotesize
    \setlength{\tabcolsep}{2.0pt}
    \renewcommand{\arraystretch}{1}
    \caption{\textbf{Random examples from \acronym{}}. }
    \begin{tabular}{@{}lp{3cm}}
    \toprule
    \textbf{Image} & \textbf{AdVQA}\\\midrule
    \raisebox{-.5\height}{\includegraphics[width=5cm]{images/r10_censored.png}} &
    \hbox{\textbf{Q}: Which hands does he have bracelets on?}
    \hbox{\textbf{Processed Answers}: both (count: 9)}
    \hbox{\textbf{Raw Answers}: 'both', 'both hands', 'both', 'both'}
    \hbox{'both', 'both', 'both', 'both', 'both', 'both'}\\
    \raisebox{-.5\height}{\includegraphics[width=5cm]{images/r9_censored.png}} &
    \hbox{\textbf{Q}:What is the baby wearing?}
    \hbox{\textbf{Processed Answers}: overalls (count: 6)}
    \hbox{\textbf{Raw Answers}: 'shirt \& overall', 'overalls', 'overalls', 'overalls'}
    \hbox{'jumper', 'overalls', 'coverals ', 'dungerees', 'overalls', 'overalls'}
    \end{tabular}
    \label{tab:random_3}
\end{table*}
\begin{table*}[h]
    \centering
    \footnotesize
    \setlength{\tabcolsep}{2.0pt}
    \renewcommand{\arraystretch}{1}
    
    \begin{tabular}{@{}lp{3cm}}
    \toprule
    \textbf{Image} & \textbf{AdVQA}\\\midrule
    \raisebox{-.5\height}{\includegraphics[width=5cm]{images/r1_censored.png}} &
    \hbox{\textbf{Q}: How many people can be seen in the room?}
    \hbox{\textbf{Processed Answers}: 6 (count: 5), 7 (count: 3)}
    \hbox{\textbf{Raw Answers}: '7', '5', '6', '8', '6', '7', '6', '6', '7', '6'}
    \\
    \raisebox{-.5\height}{\includegraphics[width=5cm]{images/r2.jpeg}} &
    \hbox{\textbf{Q}: What is on the stovetop?}
    \hbox{\textbf{Processed Answers}: kettle (count: 4)}
    \hbox{\textbf{Raw Answers}: 'tea kettle', 'kettle', 'teapot', 'teapot', 'right'}
    \hbox{'tea kettle', 'kettle', 'teapot', 'kettle', 'kettle'}
    \\
    \raisebox{-.5\height}{\includegraphics[width=5cm]{images/r3.jpeg}} &
    \hbox{\textbf{Q}: Whats does the three letters spell?}
    \hbox{\textbf{Processed Answers}: unknown (not in vocab)}
    \hbox{\textbf{Raw Answers}: 'pub', 'pub', 'pub', 'pub', 'pub'}
    \hbox{'pub', 'pub', 'pub', 'pub', 'pub'}
    \\
    \raisebox{-.5\height}{\includegraphics[width=5cm]{images/r4.jpeg}} &
    \hbox{\textbf{Q}: Are the windows all the same size?}
    \hbox{\textbf{Processed Answers}: no (count: 10)}
    \hbox{\textbf{Raw Answers}: 'no', 'no', 'no', 'no', 'no', 'no'}
    \hbox{'no', 'no', 'no', 'no'} 
    \\
    \end{tabular}
    \label{tab:random_1}
\end{table*}

\begin{table*}[t]
    \centering
    \footnotesize
    \setlength{\tabcolsep}{2.0pt}
    \renewcommand{\arraystretch}{1}
    \begin{tabular}{@{}lp{3cm}}
    \toprule
    \textbf{Image} & \textbf{AdVQA}\\\midrule
    \raisebox{-.5\height}{\includegraphics[width=5cm]{images/r5.jpeg}} &
    \hbox{\textbf{Q}: how many pieces of meat?}
    \hbox{\textbf{Processed Answers}: 2 (count: 7), 1 (count: 3)}
    \hbox{\textbf{Raw Answers}: '2', '2', '2', '2', '2', '2', '1', '1', '1', '2'} \\
    \raisebox{-.5\height}{\includegraphics[width=5cm]{images/r6.jpeg}} &
    \hbox{\textbf{Q}: Which is the largest window?}
    \hbox{\textbf{Processed Answers}: middle (count: 3)}
    \hbox{\textbf{Raw Answers}: 'center', 'bottom middle', 'bottom middle', } 
    \hbox{'middle', 'middle', 'bottom middle', 'middle lower '} 
    \hbox{'where a cat sits', 'middle one', 'the middle'} \\
    \raisebox{-.5\height}{\includegraphics[width=5cm]{images/r7.jpeg}} &
    \hbox{\textbf{Q}: What color is the checkerboard background?}
    \hbox{\textbf{Processed Answers}: black and white (count: 5)}
    \hbox{\textbf{Raw Answers}: 'unanswerable', 'black and white', 'gray'} 
    \hbox{'black and white', 'black', 'black white', 'black and white'} 
    \hbox{'black and white', 'black white', 'black and white'} \\    
    \raisebox{-.5\height}{\includegraphics[width=5cm]{images/r8.jpeg}} &
    \hbox{\textbf{Q}: What does it say on the side of the boat closest in the foreground?}
    \hbox{\textbf{Processed Answers}: unknown (not in vocab)}
    \hbox{\textbf{Raw Answers}: 'sanssouci', 'sanssouci', 'sanssouci'} 
    \hbox{'sanssouci', 'sanssouci', 'sanssouci', 'sanssouci', 'sanssouci'} 
    \hbox{'sanssouci', 'sanssouci'} 
    \end{tabular}
    \label{tab:random_2}
    
\end{table*}

\end{document}